\documentclass[10pt,twocolumn,letterpaper]{article}

\usepackage[pagenumbers]{cvpr}      

\usepackage[dvipsnames]{xcolor}

\usepackage{booktabs}
\usepackage{amsmath}
\usepackage{amssymb}
\usepackage{amsfonts}
\usepackage{pifont}
\usepackage{graphicx}
\usepackage{lipsum} 
\usepackage{caption}
\usepackage{esvect}
\usepackage{placeins}

\newcommand{\tick}{\ding{51}}%

\def \modelname {\mbox{RPM}}
\def \datasetname {\mbox{GORP}}
\def \aujst {$\text{AUJ}_{\text{S-T}}$}
\def \aujts {$\text{AUJ}_{\text{T-S}}$}
\def \pjst {$\text{PJ}_{\text{S-T}}$}
\def \pjts {$\text{PJ}_{\text{T-S}}$}
\def \supp {supp. material} 

\usepackage{algorithm}
\usepackage{algorithmic}

\definecolor{cvprblue}{rgb}{0.21,0.49,0.74}
\usepackage[pagebackref,breaklinks,colorlinks,citecolor=cvprblue]{hyperref}

\usepackage[capitalize]{cleveref}
\crefname{section}{Sec.}{Secs.}
\Crefname{section}{Section}{Sections}
\Crefname{table}{Table}{Tables}
\crefname{table}{Tab.}{Tabs.}

\def\paperID{780} 
\def\confName{CVPR}
\def\confYear{2025}

\title{From Sparse Signal to Smooth Motion: Real-Time Motion Generation with Rolling Prediction Models}

\author{German Barquero$^{1,2,3}$,  Nadine Bertsch$^{1}$,  Manojkumar Marramreddy$^{1}$,  Carlos Chacón$^{1}$,  Filippo Arcadu$^{1}$ \\ 
Ferran Rigual$^{1}$,  Nicky Sijia He$^{1}$,  Cristina Palmero$^{2,4}$,  Sergio Escalera$^{2,3}$,  Yuting Ye$^{1}$,  Robin Kips$^{1}$ \vspace{0.2cm}\\
$^{1}$Meta Reality Labs, $^{2}$Universitat de Barcelona, $^{3}$Computer Vision Center, $^{4}$King's College London \\
\small\url{https://barquerogerman.github.io/RPM/} \vspace{-0.6cm}
}

\begin{document}

\renewcommand\figureautorefname{Fig.}
\renewcommand\equationautorefname{Eq.}
\renewcommand\tableautorefname{Tab.}
\renewcommand\subsectionautorefname{Sec.}
\renewcommand\sectionautorefname{Sec.}

\twocolumn[{%
\renewcommand\twocolumn[1][]{#1}%
\maketitle

\begin{center}
    \centering
    \captionsetup{type=figure}
    \includegraphics[width=\textwidth]{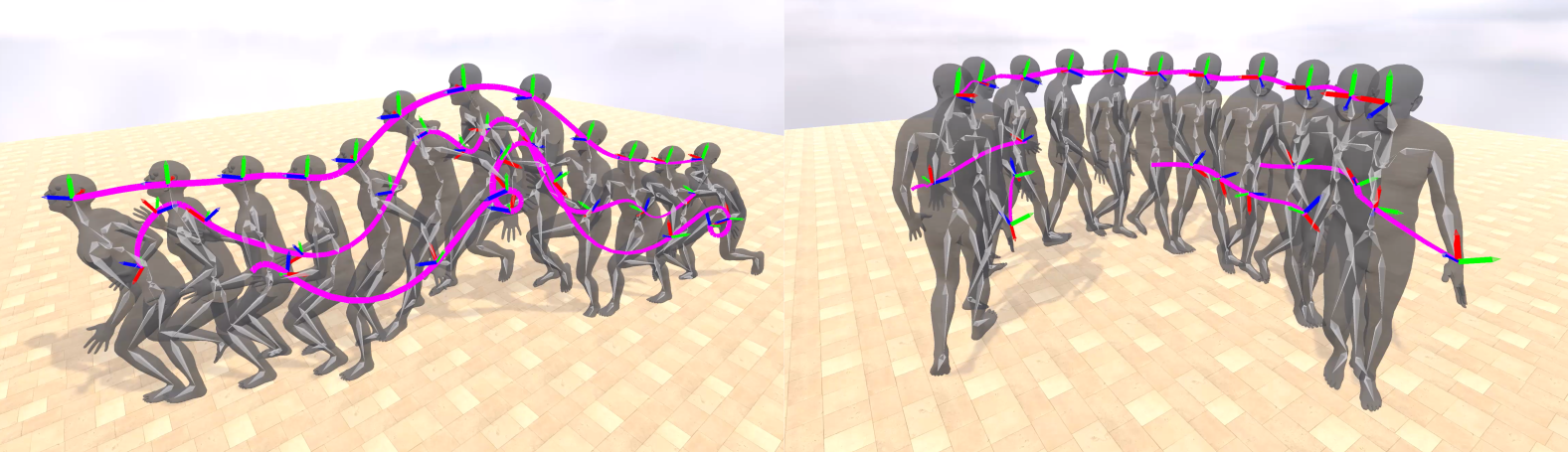}
    \vspace{-6mm}
    \captionof{figure}{We introduce Rolling Prediction Model, an 
    approach that generates smooth and realistic full-body human motion in the two of the most common XR sensing signals: hand controllers, in which the tracking signal is always available (left), and hand tracking, in which the tracking signal is noisy and might be lost for long periods of time (right). Tracking input trajectories are shown as \textbf{\textcolor{magenta}{magenta}} lines.
    }
    \label{fig:teaser}
\end{center}%
}]

\def\p{\textbf{p}}
\def\P{\mathcal{P}}
\def\x{\textbf{x}}
\def\X{\mathcal{X}}
\def\c{\textbf{c}}
\def\C{\mathcal{C}}

\definecolor{DarkCyan}{rgb}{0.4, 0.85, 0.8}
\newcommand{\RK}[1]{{\color{DarkCyan}{kipsr: #1}}}

\begin{abstract}
In extended reality (XR), generating full-body motion of the users is important to understand their actions, drive their virtual avatars for social interaction, and convey a realistic sense of presence. While prior works focused on spatially sparse and always-on input signals from motion controllers, many XR applications opt for vision-based hand tracking for reduced user friction and better immersion. Compared to controllers, hand tracking signals are less accurate and can even be missing for an extended period of time.
To handle such unreliable inputs, we present Rolling Prediction Model (RPM), an online and real-time approach that generates smooth full-body motion from temporally and spatially sparse input signals. Our model generates 1) accurate motion that matches the inputs (i.e., tracking mode) and 2) plausible motion when inputs are missing (i.e., synthesis mode). More importantly, RPM generates seamless transitions from tracking to synthesis, and vice versa. To demonstrate the practical importance of handling noisy and missing inputs,
we present GORP, the first dataset of realistic sparse inputs from a commercial virtual reality (VR) headset with paired high quality body motion ground truth. GORP provides $>$14 hours of VR gameplay data from 28 people using motion controllers (spatially sparse) and hand tracking (spatially and temporally sparse). We benchmark RPM against the state of the art on both synthetic data and GORP to highlight how we can bridge the gap for real-world applications with a realistic dataset and by handling unreliable input signals.  
Our code, pretrained models, and GORP dataset are available in the project webpage.

\end{abstract}    
\vspace{-3mm}
\section{Introduction}
\label{sec:intro}

We are witnessing an explosion of extended reality (XR) applications in recent years~\cite{ashtari2020creating, smutny2019review}.
Whether in a fully immersive virtual world, or in augmented reality (AR), being aware of the users' full-body motion is critical to understand their interactions with the world and other people. 
However, the users' full-body motion can only be inferred from spatially-sparse egocentric sensing signal from Inertial Measurements Units (IMU) or SLAM.
Prior works achieved good results assuming the input signals are always available and reliable~\cite{jiang2022avatarposer, zheng2023avatarjlm, du2023agrol, castillo2023bodiffusion, dai2024hmdposer}. While this is mostly true for motion controllers (MC), it does not apply to the frequent and more user-friendly hand tracking (HT) setup. In this mode, hand occlusions are frequent, leading to missing hand signals as in~\autoref{fig:teaser}. 
During such \textit{sensing signal losses}, a good solution should still synthesize plausible full-body motion (i.e., synthesis mode). When the missing hand signal reappears (i.e., tracking mode), its position and motion dynamics may not match that of the current synthesis result. In this situation, rather than snapping the tracked hand immediately, a smooth transition with realistic movement dynamics is more desirable, as motion discontinuities break the perceived motion realism~\cite{barquero2023belfusion}, which is critical to ensure the immersion during co-presence~\cite{toothman2019impact, almeida2022telepresence}. And the level of \textit{reactiveness} may be application dependent.

Despite the importance of this problem, only few works so far explored HT scenarios in XR~\cite{aliakbarian2023hmdnemo, qian2024reliaavatar}. One of the main barrier is the lack of datasets with real sensing signals and high quality ground truth. Collecting such a dataset requires solving the tedious task of precise synchronization and calibration between sensing data from a commercial XR device and motion capture (MoCap) body ground truth.
Instead, prior works train and validate their methods using synthetic MC and HT setups from MoCap data~\cite{mahmood2019amass}. These benchmarks, however, fall short of replicating many challenges encountered in real-world usages, such as frequent tracking signal loss and noisy hand-tracking inputs. As a result, models trained on synthetic data often underperform when exposed to real tracking inputs~\cite{dai2024hmdposer}, highlighting the need for new datasets capturing real XR scenarios.

In this work, we present Rolling Prediction Model (\modelname{}), a new architecture designed to address the challenges of real XR scenarios. Our main contributions are:

\begin{itemize}
    \item We introduce \modelname{}, a novel autoregressive model for online and real-time human motion generation from both temporally and spatially sparse tracking inputs. At every timestep, it progressively refines its prediction of the future full-body motion based on newly available input signals. By reformulating the generation process into a progressive refinement process, our approach becomes robust to long periods of hand-tracking losses and is able to generate smooth transitions when tracking recovers. Thanks to our Prediction Consistency Anchor Function (PCAF), we can control the trade-off between smooth motion dynamics and reactiveness to input signals.
    \item We present \datasetname{}, the first dataset of paired real VR tracking data with MoCap ground truth. The tracking data is obtained from hand tracking and motion controllers of a commercial VR device, revealing practical real-world challenges. It exposes new challenges for research in full-body motion tracking from sparse input.
\end{itemize}

\section{Related work}
\label{sec:related_work}

\textbf{Spatially sparse observations.} Research on this domain has focused on generating full-body motion from spatially sparse observations such as MoCap markers~\cite{ghorbani2021soma, loper2014mosh}, IMU signals~\cite{yi2022pip, yi2021transpose, van2024diffusionposer, mollyn2023imuposer, jiang2022tip}, or head-mounted cameras~\cite{jiang2022avatarposer, zheng2023avatarjlm, castillo2023bodiffusion, dai2024hmdposer, aliakbarian2023hmdnemo, qian2024reliaavatar, du2023agrol, di2023dap, dong2024mmd, starke2024codebook, feng2024sage, guzov2024hmd, jiang2025egoposer, yi2024estimating}. 
Most of these works rely on current and past tracking inputs to make individual pose predictions. However, because sparse tracking inputs are inherently ambiguous (i.e.,  multiple plausible full-body motions)
, these methods often produce jittery motion that lacks temporal coherence. To ensure motion smoothness and coherence, some of them incorporate autoregressive modules such as recurrent neural networks (RNNs)~\cite{dai2024hmdposer, qian2024reliaavatar}, or use diffusion models
~\cite{van2024diffusionposer, du2023agrol, castillo2023bodiffusion}. However, the former are prone to degenerate when errors accumulate and the latter are slow due to long denoising processes. 
Most of these works consider the scenario where tracking signal is always available and accurate, and simulate tracking inputs for MoCap datasets using the ground truth position of the head and wrists. As a result, their performance degradates when using real tracking inputs~\cite{dai2024hmdposer}.

\textbf{Temporally sparse observations.} When the tracking signal is also temporally sparse such as when using hand tracking modules, we need to 
promote the temporal motion coherence even when tracking signal is missing. Most frequent techniques include the use of a state that carries on the information from past iterations in an autoregressive fashion. HMD-NeMo~\cite{aliakbarian2023hmdnemo} was the first work to acknowledge this problem. At every timestep, their method predicts the hands latent features for the next timestep. In case a hand is missing, they can fill in the anticipated latent features. EgoPoser~\cite{jiang2025egoposer} simulates the field of view during training and presents a global motion decomposition technique. ReliaAvatar~\cite{qian2024reliaavatar} also proposed a method to fill long tracking inputs gaps. However, these methods overlook a central problem in this task: synthesis-to-tracking transitioning. When trained in a supervised setup, they learn to break the continuity of their internal state to match the new tracking input, thus destroying the motion smoothness and realism.

\textbf{Realistic motion transitioning.} Generating motion transitions is challenging. The number of possible consecutive motions is large, and the time it takes to realistically transition between actions is variable.
Some works have explored diffusion-based motion priors to make these transitions emerge \cite{shafir2023priormdm, barquero2024flowmdm, petrovich2024multitrack}. However, these are slow at inference time and are not compatible with an online setting. In our scenario, we only have access to the history of previously generated motion. 
After a long period of a hand-tracking signal loss, we need to realistically match again the recovered tracking signal. However, very frequently, the synthesized motion might show the user's arm far from the real user's hand. In these cases, what is the appropriate trade-off between the realism of the transitioning motion and the system reactiveness to match the tracking signal? This trade-off might be application dependent. For example, while social applications usually require slower transitions to avoid breaking the immersion, some games would rather trade off smoothness in favor of reactive motion generation. Our proposed model allows such control.

\begin{figure}
    \centering
    \includegraphics[width=\linewidth]{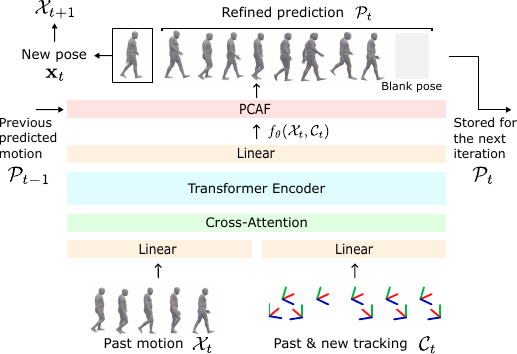}
\vspace{-6mm}
    \caption{Our \modelname{} is conditioned on the past generated motion and the past and present tracking inputs. It outputs the predicted motion, which is fed to the PCAF module in the next iteration.}
    \label{fig:arch}
\vspace{-3mm}
\end{figure}

\section{Methodology}
\label{sec:methodology}


\textbf{Problem formulation.} Our objective is to generate a sequence of human motion represented as $S{=}\{\x_1, \dots, \x_N\}$, where each $\x_t$ consists of a feature vector defining the body pose at time $t\in [1, N]$.
The generation needs to be online, that is, the pose $\x_t$ can be generated only using the sequence of past and current tracking inputs, $C_{1:t}{=}\{\c_1, \dots, \c_t\}$, and the previously generated body poses, $S_{1:t{-}1}{=}\{\x_1, \dots, \x_{t-1}\}$.

\subsection{Rolling Prediction Models}
\label{sec:method_rdm}

When generating motion from dynamic and ambiguous sparse tracking inputs, we often observe sudden drifts on the generated motion. These are mainly caused by the arrival of new tracking inputs, which update the network belief on the user pose. This problem worsens when tracking signal becomes temporally sparse. Distance-based metrics force the network to always match the new tracking signal, even if this means breaking the motion smoothness. Some works try to alleviate such problem by incorporating smooth losses such as in \cite{dai2024hmdposer}. However, systematically promoting smooth motion induces the network to remove high-frequency motions that might be of interest, such as waving.

In this work, we propose an alternative way to approach this problem: the Rolling Prediction Model. Our hypothesis is that the generation of motion can be split into a sequential process where we first generate a rough and oversmoothed motion (e.g., low-frequency motion), and then add the details (e.g., high-frequency motion), as in diffusion models~\cite{yang2023diffusion, ruhe2024rolling, zhang2023tedi, tevet2023human, zhang2024motiondiffuse}. Similarly, in the human motion prediction field, numerous studies have observed that neural networks tend to predict detailed high-frequency motion in the short term, and are only able to predict low-frequency motion in the long term~\cite{barquero2022comparison, barquero2022didn, palmero2022chalearn, martinez2017human, lyu20223d, rudenko2020human, barquero2023belfusion, xu2022diverse, lyu20223d}. This is due to the inherent uncertainty of human behavior. \modelname{} combines both observations into a single paradigm: generating motion by progressively refining a prediction of the following motion as time advances and new tracking input is received. 

Most prior works that deal with this specific problem use the past and present tracking inputs to generate the full-body pose at the current timestep $t$, that is, $f_\theta (\C_t) = \x_t$, where $\C_t{=}\{\c_{t-I}, \dots, \c_{t}\}$ are the $I{+}1$ past tracking inputs. Instead, we generate, or predict, the following $W$ poses:

\vspace{-3mm}
\begin{equation}
    f_{\theta}(\X_t, \C_t){=}\{\hat{\x}_{t}, \hat{\x}_{t+1}, \dots, \hat{\x}_{t+W}\},
\end{equation}

\noindent where $\X_t{=}\{\x_{t-M}, \dots, \x_{t-1}\}$ are the $M$ past generated poses kept as context, and $f_\theta$ is our network.
Following the rationale explained before, this predicted motion can be reasonably accurate in the short-term and will be only a rough approximation in the long-term. Therefore, in the next timestep, $t{+}1$, 
the previously predicted pose $\hat{\x}_{t+1}$ will already be a good approximation of the new pose that needs to be generated. More specifically, if we could re-use the previous predicted future poses, the correction predicted by the network would be small in the short term and large in the long term. 
If such functioning were also held when a tracking input is suddenly lost or, on the contrary, recovered from a loss, the predicted motion would be slowly refined to incorporate such new tracking input information (or the lack of it) and, therefore, the resulting transition would be smooth. Unfortunately, this behavior does not happen naturally when training $f_\theta$ supervisedly, as any distance-based loss would break the smoothness of the motion in favor of instantly matching the newly acquired tracking input. 
Instead, we propose to inject such inductive bias by design with our novel Prediction Consistency Anchor Function (PCAF), which reparameterizes the output of the network at timestep $t$ as follows:

\vspace{-5mm}
\begin{equation}
\label{eq:dcaf}
\P_{t} = \P_{t-1} + U \cdot \tanh (f_\theta (\X_t , \C_t) - \P_{t-1}),
\end{equation}

\noindent where $U{=}(u_1, u_2, \dots, u_W) {\in} [0, 1]^W$ represents the uncertainty along the future predicted poses, and $\P_{t-1}$ is the motion prediction from the previous timestep, see \autoref{fig:arch}. After every iteration, we move $\P_t$ one timestep forward into the future by removing the first pose $\x_t$ and appending a blank pose at the end of it. Note that under long signal losses, the historical sequence of tracking inputs $\mathcal{C}_t$ empties and the model relies only on the past predictions $\mathcal{X}_t$.
By combining \modelname{} with PCAF, we achieve two things. First, we force the network to break down the generation of a pose into $W$ rolling steps, in which major changes to a pose can only be introduced when such pose is still distant from the present (i.e., highest uncertainty). When the predicted pose is already close to the present, only minor adjustements can be injected, ensuring that the recent updates from the tracking inputs are incorporated. As a result, the network learns to plan ahead in time, and leverages such capabilities to improve in the generation task.
Second, the reactiveness of the network to the tracking input can be controlled now by either modifying the uncertainty schedule $U$, or predicting shorter- or longer-term motion (i.e., smaller or larger $W$). 

There are a few things to consider regarding the PCAF definition. First, it might look counterintuitive that without access to the previous prediction, the network can \textit{blindly} correct it. However, this is not the case here. The network still has access to the past motion context $\X_{t}$ which serves as an explicit motion state and, under the deterministic human motion prediction umbrella, is a strong proxy for the previous prediction, $\P_{t-1}$. As a result, the network is able to implicitly recover the previous prediction, and inject the higher-frequency details that the access to the new tracking input unlocks (further discussed in \supp{}). 
Second, the choice of a hyperbolic tangent is made for optimization purposes. When the generated motion is severely misaligned with the tracking input during training time, the magnitude of the gradient for the short-term future prediction will become very small (i.e., high absolute difference between current and previous prediction), and therefore focus on improving the long-term prediction capabilities. 
Finally, note that PCAF does not bound the magnitude of the predicted motion, but the correction with respect to the previous prediction. Therefore, it can still generate highly dynamic motion when needed, as we show in \autoref{sec:qualitative_results}.

\subsection{Free-Running Training Strategy}
\label{sec:method_fr}

\begin{figure}
    \centering
    \includegraphics[width=\linewidth]{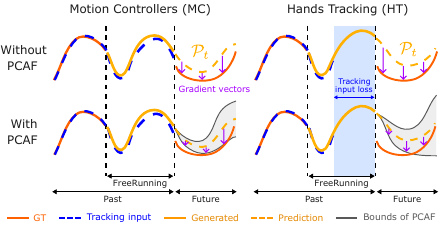}
    \vspace{-6mm}
    \caption{While in free-running, the tracking signal and the generated motion might misalign. When applying a distance-based loss, the gradient pushes to correct the predicted motion and immediately match the tracking signal. This makes the model generate jittery motion and abrupt transitions after tracking input losses (top row). PCAF forces the magnitude of the correction to be within the bounds of the PCAF uncertainty (bottom row).}
    \label{fig:free_running_dcaf}
\vspace{-3mm}
\end{figure}

\def\fr{\text{FR}}

When generating motion from sparse tracking inputs in an online way, the arrival of new signal might reduce or increase the uncertainty of the corresponding full-body pose. As a result, the ongoing motion at time $t$ might need to be corrected at time $t+1$. These complex dynamics between tracking signal and generated motion might lead to misalignments at inference time between $\X_t$ and $\C_t$ that never happened during training, where $\X_t$ is always ground truth past poses. As a result, as soon as there is a misalignment between $\X_t$ and $\C_t$, the generated motion progressively degenerates. To improve the model stability and robustness to its own mistakes, we integrate a free-running stage at the beginning of every training iteration, as described in Alg.~\ref{alg:consistentrd}. For this, we simulate the generation of $fr\sim \mathcal{U}(0, \text{FR})$ poses. During this period, the context of past generated motion $\X_{t{+}fr}$ is populated with poses predicted by the model, being exposed to its own errors during training. As a result, the network learns to correct these misalignments between tracking input and generated motion that will also be encountered during inference. Thanks to PCAF, such corrections are subtle and progressive, as shown in \autoref{fig:free_running_dcaf}.

\begin{algorithm}[t!]
\small
  \caption{Training of Rolling Prediction Model.}
  \label{alg:consistentrd}
  \begin{algorithmic}[1]
    \REQUIRE $f_{\theta}$, $C{=}\{\c_{1}, \dots, \c_{N}\}$, $S_{\text{GT}}{=}\{\x_{1}, \dots, \x_{N}\}$
    \STATE Sample $t \sim \mathcal{U}(\max (I,M), N{-}\text{FR}{-}W)$
    \STATE Get $\C:=C[t{-}I: t]$,  $\X:=S_\text{GT}[t{-}M:t{-}1]$
    \STATE Initialize $\P :=  S_\text{GT}[t:t{+}W]$
    \STATE \# ========= Begin free-running =========
    \STATE Sample $fr \sim \mathcal{U}(0, \text{FR})$
    \STATE \textbf{with} torch.no\_grad:
    \begin{ALC@g}
    \FOR{$i{=}1$ to $fr$}
        \STATE Prediction step: $f_{\theta}(\X, \C)$
        \STATE $\P:=$ PCAF reparameterization as in \autoref{eq:dcaf}
        \STATE Remove first element from $\C$ and $\X$
        \STATE Pop first pose from $\P$ and append it to $\X$
        \STATE Append new input $\c_{t+i}$ to $\C$
        \STATE Set last pose of $\P$ to zeros (blank pose)
        
    \ENDFOR
    \end{ALC@g}
    \STATE \# ========= End of free-running =========
    \STATE \textbf{with} torch.grad:
    \begin{ALC@g}
        \STATE Prediction step: $f_{\theta}(\X, \C)$
        \STATE $\P:=$ PCAF reparameterization as in \autoref{eq:dcaf}
        \STATE Get $\P_{\text{GT}}:=S_{\text{GT}}[t{+}fr:t{+}fr{+}W]$
        \STATE Update $\theta$ with $\mathcal{L}(\P, \P_{\text{GT}})$ as in \autoref{eq_loss}
    \end{ALC@g}
  \end{algorithmic}
\end{algorithm}

\subsection{Optimization}
We optimize the model with the following loss:

\def\orient{\text{ori}}
\def\relrot{\text{rot}}
\def\pos{\text{pos}}
\def\velorient{\vv{\text{ori}}}
\def\velrot{\vv{\text{rot}}}
\def\velpos{\vv{\text{pos}}}

\def\lorient{\mathcal{L}_{\orient}}
\def\lrelrot{\mathcal{L}_{\relrot{}}}
\def\lpos{\mathcal{L}_{\pos}}
\def\lvelorient{\mathcal{L}_{\velorient}}
\def\lvelrot{\mathcal{L}_{\velrot}}
\def\lvelpos{\mathcal{L}_{\velpos}}

\vspace{-5mm}
\begin{equation}
\label{eq_loss}
\begin{split}
\mathcal{L} = \lambda_{\orient}\lorient + 
\lambda_{\relrot{}}\lrelrot + 
\lambda_{\pos}\lpos +  \\
\lambda_{\velorient}\lvelorient +
\lambda_{\velrot}\lvelrot +
\lambda_{\velpos}\lvelpos,
\end{split}
\end{equation}

where $\lorient$ is the global orientation loss, $\lrelrot$ is the relative rotations loss, $\lpos$ is the joints position loss, and $\lvelorient$, $\lvelrot$, and $\lvelpos$ are the same losses applied to their velocities. Joints positions are computed with a forward kinematics model~\cite{pavlakos2019expressive} using the ground truth shape parameters of the individual. All losses are computed as the differences of absolute values (i.e., L1 distance) between the poses in the last predicted window $\P_{t+fr}$ after the $fr$ free-running steps and the ground truth value in that motion window. 
Note that the loss is not backpropagated through time along the free-running process to save memory during training. 

\section{Experimental results}
\label{sec:experiments}
In this section, we compare \modelname{} to the state of the art on synthetic benchmarks and ablate each of our contributions. Then, we present the most comprehensive evaluation up to date on real-world data using the new \datasetname{} dataset.

\textbf{Datasets.}
We replicate prior works' synthetic benchmarks based on AMASS~\cite{mahmood2019amass}: an intra-dataset split (A-P1, 60 FPS) \cite{AMASS_CMU, AMASS_HDM05, AMASS_BMLrub} and a cross-dataset split (A-P2, 30 FPS) \cite{AMASS_ACCAD, AMASS_BMLrub, AMASS_BMLmovi, AMASS_CMU, AMASS_KIT-CNRS-EKUT-WEIZMANN, AMASS_EyesJapanDataset, AMASS_HumanEva, AMASS_KIT-CNRS-EKUT-WEIZMANN-3, AMASS_MoSh, AMASS_HDM05, AMASS_PosePrior, AMASS_SFU, AMASS_TotalCapture}. 
These benchmarks are built by taking the ground truth head and wrist joints as spatially sparse tracking inputs (MC). We also build their synthetic spatially and temporally sparse versions (HT) by simulating random tracking signal losses lasting from 0.5s to 2s. Note that, instead of assuming an average male body shape for all sequences as in \cite{jiang2022avatarposer, zheng2023avatarjlm, du2023agrol}, we use non-gendered SMPL-X meshes~\cite{pavlakos2019expressive} with their corresponding shape coefficients. We argue that this is a more realistic setup in which evaluation also accounts for how well the model understands the user dimensions. 
Still, A-P1 and A-P2 do not contain the challenges of real MC/HT scenarios. To fill this gap, we evaluate on \datasetname{}, our new MoCap dataset with real inputs of $>$14 hours with 28 participants playing in VR on real MC and HT setups (30 FPS).

\textbf{Evaluation.} Similarly to~\cite{jiang2022avatarposer, zheng2023avatarjlm, du2023agrol, dai2024hmdposer}, we use the MPJRE (Mean Per Joint Rotation Error, in degrees), the MPJPE (Mean Per Joint Position Error, in cm), and the MPJVE (Mean Per Joint Velocity Error, in cm/s). To measure the smoothness of the motion, we compute the Jitter ($10^2 m/s^3$) as in \cite{du2023agrol}. To measure the smoothness during transitions, we use the Peak Jerk (PJ) and Area Under the Jerk (AUJ) as proposed in \cite{barquero2024flowmdm}. More specifically, we compute the PJ and AUJ for a 1-second period of motion following the two types of transitions: from tracking to synthesis modes (T-S), and vice versa (S-T). To compute the joints positions, we apply forward-kinematics with the SMPL-X ground truth body shape parameters, assuming that, in a real deployment, these would be available from a prior calibration step. Our evaluation skips an initial 1-second padding in all sequences to allow the methods to warm up.

\textbf{Baselines.} We compare \modelname{} with AvatarPoser~\cite{jiang2022avatarposer}, AGRoL~\cite{du2023agrol}, AvatarJLM~\cite{zheng2023avatarjlm}, SAGE~\cite{feng2024sage}, EgoPoser~\cite{jiang2025egoposer} 
and HMD-Poser~\cite{dai2024hmdposer}. 
We randomly drop periods of variable length of each hand-tracking input to simulate tracking signal losses during training. The shape head was removed from HMD-Poser. The ground truth shape was used for all methods with forward-kinematics-based losses.


\textbf{Implementation details.}
Our motion representation consists of a vector of 6D rotations~\cite{zhou2019continuity} for the global body orientation, and the 6D rotation of all bone rotations.
. Tracking signal contains the 6-DOF (position + rotation) from the head and wrists as in \cite{jiang2022avatarposer, du2023agrol}. The world trajectory is given by the head position. 
FR was set to 60/30 frames for A-P1/A-P2 and \datasetname{} (i.e., 1s). The uncertainty function $U$ was set to a cosine-based function, akin to the one used in diffusion models~\cite{nichol2021improved} (see \supp{}). Tracking input and motion contexts are set to $I{=}M{=}10$ frames. For our reactive/smooth \modelname{}, $W$ is set to 10/20 frames for A-P1 and 5/10 frames for A-P2 and \datasetname{} (i.e., 0.16/0.33s). See more implementation details in \supp{}. The model is trained on a single NVIDIA A100 in 12h and, at inference, it runs in real time (206.5 FPS).

\subsection{Quantitative evaluation on synthetic data}

\begin{table*}[t!]
    \centering
    \footnotesize
    \addtolength{\tabcolsep}{-0.2em}
    \begin{tabular}{lllll|llllllll}
    \toprule
    & \multicolumn{4}{c}{Motion Controllers (MC)} & \multicolumn{8}{c}{Hand Tracking (HT)}\\
    Model & MPJRE & MPJPE & MPJVE & Jitter & MPJRE & MPJPE & MPJVE & Jitter & \pjts{} & \aujts{} & \pjst{} & \aujst{} \\
    \midrule
    AvatarPoser & $4.49$ & $5.84$ & $31.21$ & $16.09$  & $5.62$ & $8.38$ & $44.26$ & $26.10$ & $89.87$ & $2215.07$ & $86.93$ & $2133.20$ \\
    AGRoL & $4.22$ & $5.93$ & $107.21$ & $112.60$ & $5.62$ & $8.61$ & $172.82$ & $181.54$ & $1274.35$ & $34120.55$ & $1347.66$ & $29707.36$  \\
    EgoPoser & $3.40$ & $4.08$ & $27.10$ & $15.40$ & $4.61$ & $6.29$ & $42.91$ & $29.15$ & $461.18$ & $3356.38$ & $567.42$ & $3773.29$ \\
    SAGE & $3.13$ & $3.46$ & $24.17$ & $10.91$ & $4.21$ & $5.50$ & $46.38$ & $33.55$ & $1056.33$ & $6337.79$ & $1073.65$ & $4131.33$ \\
    AvatarJLM  & $3.39$ & $3.17$ & $\textbf{16.56}$ & $5.17$  & $4.18$ & $4.59$ & $27.30$ & $12.79$ & $114.36$ & $811.13$ & $901.75$ & $2008.13$\\
    HMD-Poser & $\textbf{2.77}$ & $\textbf{3.08}$ & $17.41$ & $5.96$ & $\textbf{3.34}$ & $\textbf{4.04}$ & $\textbf{22.34}$ & $7.35$ & $23.84$ & $302.58$ & $461.91$ & $1236.47$ \\
    \midrule
    \modelname - Reactive  & $3.25$ & $4.08$ & $19.21$ & $\textbf{4.21}$ & $3.82$ & $5.18$ & $22.83$ & $4.35$ & $15.28$ & $\textbf{60.51}$ & $18.98$ & $69.02$ \\
    \modelname{} - Smooth  & $3.58$ & $4.67$ & $21.47$ & $4.23$  & $3.98$ & $5.44$ & $24.04$ & $\textbf{4.29}$ & $\textbf{8.41}$ & $84.81$ & $\textbf{12.12}$ & $\textbf{50.23}$ \\
    \bottomrule
    \end{tabular}
    \vspace{-3mm}
\caption{Comparison of \modelname{} with the state of the art on A-P1. We observe how our model generates motion with less jitter, and with considerably smoother transitions (i.e., lower PJ and AUJ) from tracking to synthesis (T-S) mode, and vice versa (S-T).}
\label{tab:sota_amass1}
\vspace{-3mm}
\end{table*}

\begin{table*}[t!]
    \centering
    \footnotesize
    \addtolength{\tabcolsep}{-0.2em}
    \begin{tabular}{lllll|llllllll}
    \toprule
    & \multicolumn{4}{c}{Motion Controllers (MC)} & \multicolumn{8}{c}{Hand Tracking (HT)}\\
    Model & MPJRE & MPJPE & MPJVE & Jitter & MPJRE & MPJPE & MPJVE & Jitter & \pjts{} & \aujts{} & \pjst{} & \aujst{} \\
    \midrule
    AvatarPoser & $5.18$ & $6.65$ & $32.74$ & $2.67$ & $5.88$ & $8.38$ & $41.64$ & $4.56$ & $115.12$ & $317.11$ & $143.54$ & $380.50$ \\
    AGRoL & $5.21$ & $6.60$ & $50.11$ & $9.65$ & $6.29$ & $8.84$ & $73.01$ & $14.31$ & $90.17$ & $1080.06$ & $94.51$ & $1024.51$ \\
    EgoPoser & $5.49$ & $6.88$ & $34.19$ & $3.52$ & $6.25$ & $8.60$ & $45.61$ & $5.95$ & $71.58$ & $343.85$ & $82.28$ & $384.23$ \\
    SAGE & $5.36$ & $6.17$ & $34.62$ & $3.68$ & $6.04$ & $7.66$ & $47.63$ & $7.22$ & $147.91$ & $531.49$ & $155.55$ & $438.26$ \\
    AvatarJLM  & $5.21$ & $\textbf{5.23}$ & $\textbf{22.72}$ & $1.77$ & $5.71$ & $\textbf{6.32}$ & $30.70$ & $3.01$ & $26.45$ & $103.47$ & $115.78$ & $231.04$ \\
    HMD-Poser & $5.17$ & $5.58$ & $24.51$ & $2.15$ & $\textbf{5.60}$ & $6.42$ & $\textbf{29.35}$ & $2.52$ & $9.04$ & $55.58$ & $71.13$ & $174.80$ \\
    \midrule
    \modelname{} - Reactive  & $\textbf{5.15}$ & $6.56$ & $27.16$ & $1.55$ & $5.68$ & $7.80$ & $31.54$ & $1.64$ & $6.06$ & $11.73$ & $10.92$ & $32.81$ \\
    \modelname{} - Smooth & $5.40$ & $7.14$ & $30.36$ & $\textbf{1.39}$ & $5.92$ & $8.37$ & $34.04$ & $\textbf{1.44}$ & $\textbf{4.74}$ & $\textbf{6.43}$ & $\textbf{7.31}$ & $\textbf{11.93}$ \\
    \bottomrule
    \end{tabular}
    \vspace{-3mm}
\caption{Comparison of \modelname{} with the state of the art on A-P2.}
\label{tab:sota_amass2}
\vspace{-3mm}
\end{table*}

\begin{figure}[t!]
    \centering
    \includegraphics[width=\linewidth]{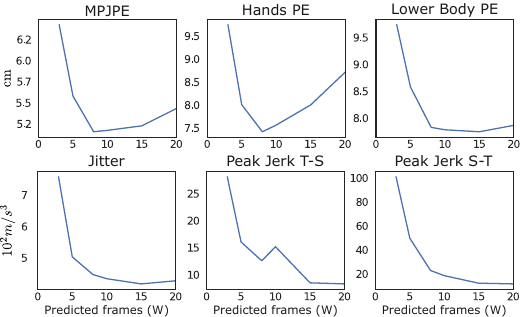}
    \vspace{-6mm}
    \caption{\textbf{Flexible reactiveness.} \modelname{} achieves the best accuracy when the prediction length is around 8 frames (or 133ms), and the smoothest results around 15 frames (or 250ms). By leveraging longer prediction windows, we can trade off smoothness for tracking reactiveness (i.e., lower jitter and peak jerk, higher hands PE).}
    \label{fig:pred_length}
\end{figure}

\begin{table}[t!]
    \centering
    \footnotesize	
    \addtolength{\tabcolsep}{-0.4em}
    \begin{tabular}{cccllllll}
    \toprule
    \modelname{} & FR & PCAF & MPJRE & MPJPE & MPJVE & Jitter & \aujts & \aujst \\
    \midrule
    \tick & & & $6.69$ & $10.59$ & $31.84$ & $5.84$ & $101.35$ & $210.63$ \\
    \tick & \tick & & $\textbf{3.79}$ & $\textbf{5.04}$ & $23.89$ & $8.32$ & $428.32$ & $799.03$ \\
    \tick& & \tick & $5.53$ & $9.07$ & $27.85$ & $\textbf{3.82}$ & $138.17$ & $116.68$ \\
    \tick & \tick & \tick & $3.82$ & $5.18$ & $\textbf{22.83}$ & $4.35$ & $\textbf{60.51}$ & $\textbf{69.02}$ \\
    \bottomrule
    \end{tabular}
    \vspace{-3mm}
\caption{\textbf{Ablation study (A-P1, HT setup).} Motion generated with \modelname{} tends to degenerate unless combined with free-running (FR). PCAF ensures \modelname{} generates smooth transitions from synthesis to tracking mode (S-T), and vice versa (T-S).}
\label{tab:ablation_contributions}
\vspace{-3mm}
\end{table}

\autoref{tab:sota_amass1} and \autoref{tab:sota_amass2} show the comparison of \modelname{} with the state of the art (SOTA) on both MC and HT scenarios, respectively. While most SOTA methods show very accurate motion generation in the MC scenario, they fail to generate smooth transitions under hand-tracking signal losses. As expected, state-less baselines such as AvatarPoser, SAGE, EgoPoser and AvatarJLM match the new hand-tracking signal as soon as it reappears, leading to abrupt transitions (i.e., very high \aujst). Even HMD-Poser, which leverages an RNN that carries on the temporal information, shows very rough transitions after tracking input losses. The generative diffusion-based baseline, AGRoL, shows high jitter when applied in an online fashion due to long jumps in the denoising chain that are needed to make it computationally efficient. By contrast, our model shows competitive accuracy in both MC/HT setups, and is the only model able to generate smooth transitions after periods of tracking input loss ($>$5x lower \aujst). By tuning the prediction length, we can increase or decrease the reactiveness of the generated motion to the tracking signal, illustrated with higher and lower PJ values, respectively, in \autoref{fig:pred_length}.

We ablate the impact of each of our contributions in \autoref{tab:ablation_contributions}. We show that, as explained in \autoref{sec:method_rdm}, vanilla \modelname{} degenerates due to misalignments between the tracking input and the synthesized motion. By incorporating the free-running stage, the model becomes robust to its own mistakes and achieves competitive accuracy. However, it still shows high jitter due to sudden corrections to the predicted motion as new tracking input is incorporated. PCAF addresses this issue by halving the jitter and reducing the \aujts{} and \aujst{} by seven and elevenfold, respectively.


\begin{table*}[t!]
    \centering
    \footnotesize
    \begin{tabular}{llll|lll|lll}
    \toprule
    & \multicolumn{6}{c}{Trained on Simulated MC} & \multicolumn{3}{|c}{Trained on Real MC} \\
    & \multicolumn{3}{c}{Evaluated on Simulated MC} & \multicolumn{3}{|c}{Evaluated on Real MC} & \multicolumn{3}{|c}{Evaluated on Real MC}\\
    Model & MPJPE & MPJVE & Jitter & MPJPE & MPJVE & Jitter & MPJPE & MPJVE & Jitter \\
    \midrule
    AvatarPoser & $5.41$ & $14.87$ & $1.48$ & $6.54$ & $16.86$ & $1.74$ & $6.49$ & $14.72$ & $1.51$ \\
    AGRoL & $5.12$ & $33.31$ & $8.78$ & $6.58$ & $38.88$ & $9.91$ & $6.14$ & $39.14$ & $10.36$ \\
    EgoPoser & $6.24$ & $16.33$ & $1.83$ & $7.51$ & $19.94$ & $2.09$ & $7.21$ & $15.00$ & $1.63$ \\
    SAGE & $5.56$ & $18.78$ & $1.86$ & $6.62$ & $21.14$ & $2.12$ & $6.49$ & $17.84$ & $1.58$ \\
    AvatarJLM & $\textbf{4.81}$ & $\textbf{9.24}$ & $0.69$ & $\textbf{5.97}$ & $\textbf{12.52}$ & $0.89$ & $\textbf{6.07}$ & $\textbf{10.79}$ & $0.70$ \\
    HMD-Poser & $5.89$ & $13.44$ & $0.94$ & $7.23$ & $19.62$ & $1.63$ & $6.84$ & $15.13$ & $0.81$ \\
    \midrule
    RPM - Reactive & $6.20$ & $11.31$ & $0.68$ & $7.36$ & $12.93$ & $0.77$ & $6.83$ & $10.93$ & $0.57$ \\
    RPM - Smooth & $6.86$ & $12.77$ & $\textbf{0.46}$ & $7.58$ & $12.92$ & $\textbf{0.47}$ & $7.36$ & $12.09$ & $\textbf{0.41}$ \\
    \bottomrule
    \end{tabular}
    \vspace{-3mm}
\caption{Comparison of \modelname{} with the state of the art on the \datasetname{} dataset when using motion controllers as tracking inputs. We observe a performance gap between training on simulated MC and training in real MC due to non-rigid position and orientation of the controller. \datasetname{} allows, for the first time, training on real controllers data, which improves the performance of all methods.}
\vspace{-2mm}
\label{tab:sota_gorp_mc}
\end{table*}

\begin{table*}[t!]
    \centering
    \footnotesize
    \addtolength{\tabcolsep}{-0.4em}
    \begin{tabular}{llllll|lllll|lllll}
    \toprule
    & \multicolumn{10}{c}{Trained on Simulated HT} & \multicolumn{5}{|c}{Trained on Real HT} \\
    & \multicolumn{5}{c}{Evaluated on Simulated HT} & \multicolumn{5}{|c}{Evaluated on Real HT} & \multicolumn{5}{|c}{Evaluated on Real HT}\\
    Model & MPJPE & MPJVE & Jitter & \aujts{} & \aujst{} & MPJPE & MPJVE & Jitter & \aujts{} & \aujst{} & MPJPE & MPJVE & Jitter & \aujts{} & \aujst{} \\
    \midrule
    AvatarPoser & $6.57$ & $6.83$ & $1.03$ & $244.48$ & $383.40$  & $7.68$ & $13.66$ & $2.93$ & $459.92$ & $1407.90$ & $7.75$ & $8.03$ & $1.23$ & $190.86$ & $396.52$\\
    AGRoL & $6.67$ & $33.50$ & $9.23$ & $974.94$ & $700.53$ & $7.52$ & $39.13$ & $10.72$ & $1214.95$ & $1121.72$ & $7.18$ & $36.74$ & $10.16$ & $978.60$ & $760.19$ \\
    EgoPoser & $7.20$ & $8.43$ & $1.44$ & $322.22$ & $395.67$ & $8.16$ & $20.03$ & $4.98$ & $703.21$ & $1079.94$ & $8.09$ & $10.22$ & $2.02$ & $318.68$ & $424.67$ \\
    SAGE & $6.08$ & $10.50$ & $1.86$ & $502.92$ & $613.40$ & $7.44$ & $19.06$ & $4.01$ & $966.41$ & $1464.03$ & $7.14$ & $11.19$ & $1.82$ & $410.99$ & $660.71$ \\
    AvatarJLM & $6.19$ & $\textbf{4.30}$ & $0.41$ & $37.22$ & $83.53$ & $\textbf{6.62}$ & $9.17$ & $1.61$ & $177.64$ & $498.39$  & $6.82$ & $5.76$ & $0.56$ & $65.22$ & $150.60$ \\
    HMD-Poser & $\textbf{6.05}$ & $5.26$ & $0.31$ & $13.10$ & $39.55$  & $7.62$ & $8.70$ & $0.70$ & $68.26$ & $201.68$ & $\textbf{6.02}$ & $7.17$ & $0.40$ & $46.18$ & $102.65$\\
    \midrule
    RPM - Reactive & $6.31$ & $4.97$ & $0.36$ & $21.88$ & $35.32$ & $7.73$ & $8.45$ & $0.81$ & $100.94$ & $239.71$ & $7.95$ & $5.64$ & $0.37$ & $20.41$ & $37.91$ \\
    RPM - Smooth & $7.31$ & $5.07$ & $\textbf{0.19}$ & $\textbf{3.93}$ & $\textbf{6.23}$ & $7.56$ & $\textbf{7.85}$ & $\textbf{0.44}$ & $\textbf{49.41}$ & $\textbf{118.12}$ & $7.57$ & $\textbf{5.41}$ & $\textbf{0.20}$ & $\textbf{4.29}$ & $\textbf{8.40}$ \\
    \bottomrule
    \end{tabular}
    \vspace{-3mm}
\caption{Comparison of \modelname{} with the state of the art on the \datasetname{} dataset when using hand-tracking signal as inputs. In this setup, the performance gap between training models on simulated and real inputs also affects the motion dynamics (MPJVE, and jitter). }
\label{tab:sota_gorp_ht}
\vspace{-3mm}
\end{table*}

\subsection{Quantitative evaluation on real VR data}
\label{sec:results_vr_data}

\textbf{Novel dataset GORP.} To evaluate our algorithm's performance under realistic scenarios, we collected a dataset of realistic VR gameplay data from Meta Quest 3~\cite{quest3}, with synchronized and calibrated ground truth body motion obtained through the Optitrack~\cite{optitrack} motion capture system (see \supp{}). 
We develop highly accurate proprietary solutions to synchronize data from the two systems and with less than one millimeter calibration error. Our 28 participants played a controller-based and a hand-tracking based game for 15 minutes each (a total of $>$14 hours). We chose games that encourage a wide range of hand and arm movements to cover a diverse set of realistic activities. They stretch the computer vision based tracking algorithms in Meta Quest 3 to reveal tracking failures when the controllers or hands are outside the headset cameras' field of view. Our dataset is the first to provide real tracking signals and high quality ground truth body motion. It exposes the field to difficult practical issues inexistent in synthetic data.

\textbf{Quantitative evaluation on GORP.} \datasetname{} enables us to study the gap between synthetic and real benchmarks. For this, we train all baselines and our method on synthetic MC and HT setups generated with \datasetname{}, as we did with A-P1 and A-P2. Then, we evaluate these methods on 1) simulated MC/HT tracking inputs, using the same segments of hand-tracking losses from the HT real inputs, and 2) real tracking input signal. As shown in \autoref{tab:sota_gorp_mc}, the MPJPE of all methods increase by around 20\% in the evaluation with real MC input, likely caused by the non-rigid position of the controllers, which can vary across subjects and along a sequence. In HT mode, see \autoref{tab:sota_gorp_ht}, the frequent tracking signal losses and the noisy hand-tracking detections impact the model's accuracy in a similar way. The motion dynamics are also affected, generating up to x4 more jittery motion with up to x2 more MPJVE (AvatarJLM). Finally, we re-train all models on real tracking inputs, which improves the accuracy and the smoothness metrics for all of them. However, despite these gains, the performance still lags behind that achieved when training and evaluating on synthetic data. This is probably due to the additional challenges of real-world input that synthetic benchmarks don't capture. 
This disparity highlights the need for more realistic benchmarks like \datasetname{}, which can help bridge the gap between synthetic and real-world performance.


\subsection{Qualitative results}
\label{sec:qualitative_results}

\begin{figure}
    \centering
    \includegraphics[width=\linewidth]{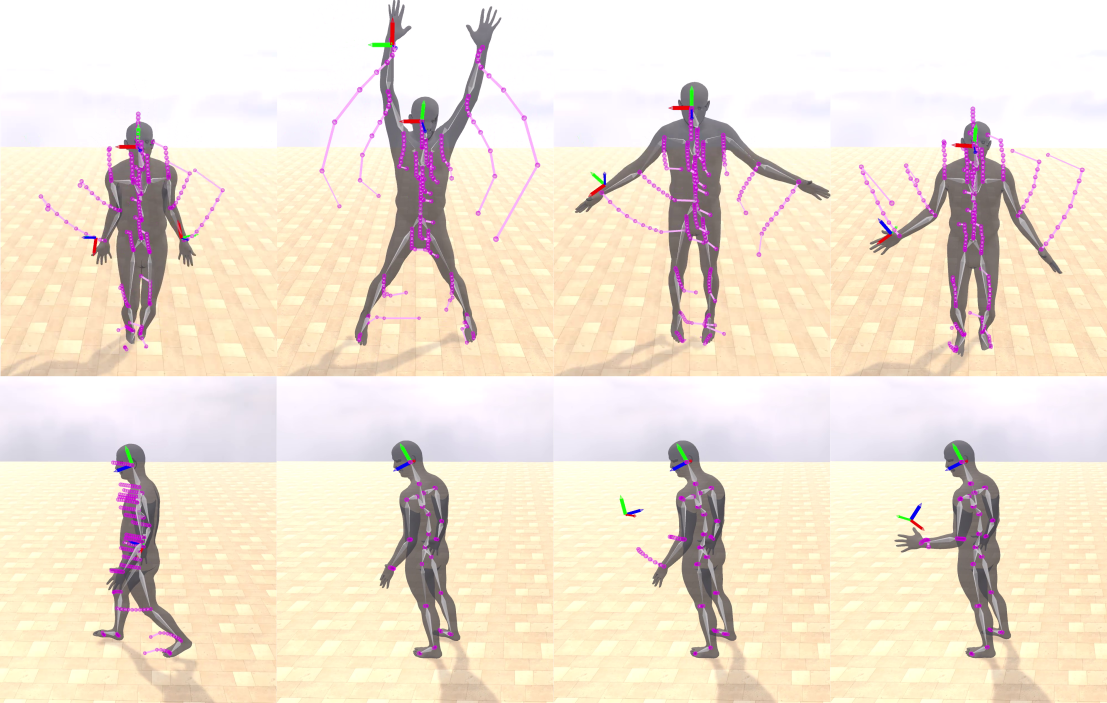}
    \vspace{-6mm}
    \caption{\textbf{Trajectory prediction.} \modelname{} decomposes the generation of motion into a progressive refinement of the predicted $W$ next poses, shown above as \textcolor{magenta}{\textbf{magenta}} dots, connected by lines. On top, we observe how \modelname{} can predict fast dynamic motion and generate expressive and realistic motion, even during tracking signal losses. Below, we show how \modelname{} generates a smooth transition when recovering from a hand-tracking loss (left hand). }
    \label{fig:trajectory_prediction}
\vspace{-3mm}
\end{figure}

\begin{figure*}
    \centering
    \includegraphics[width=\linewidth]{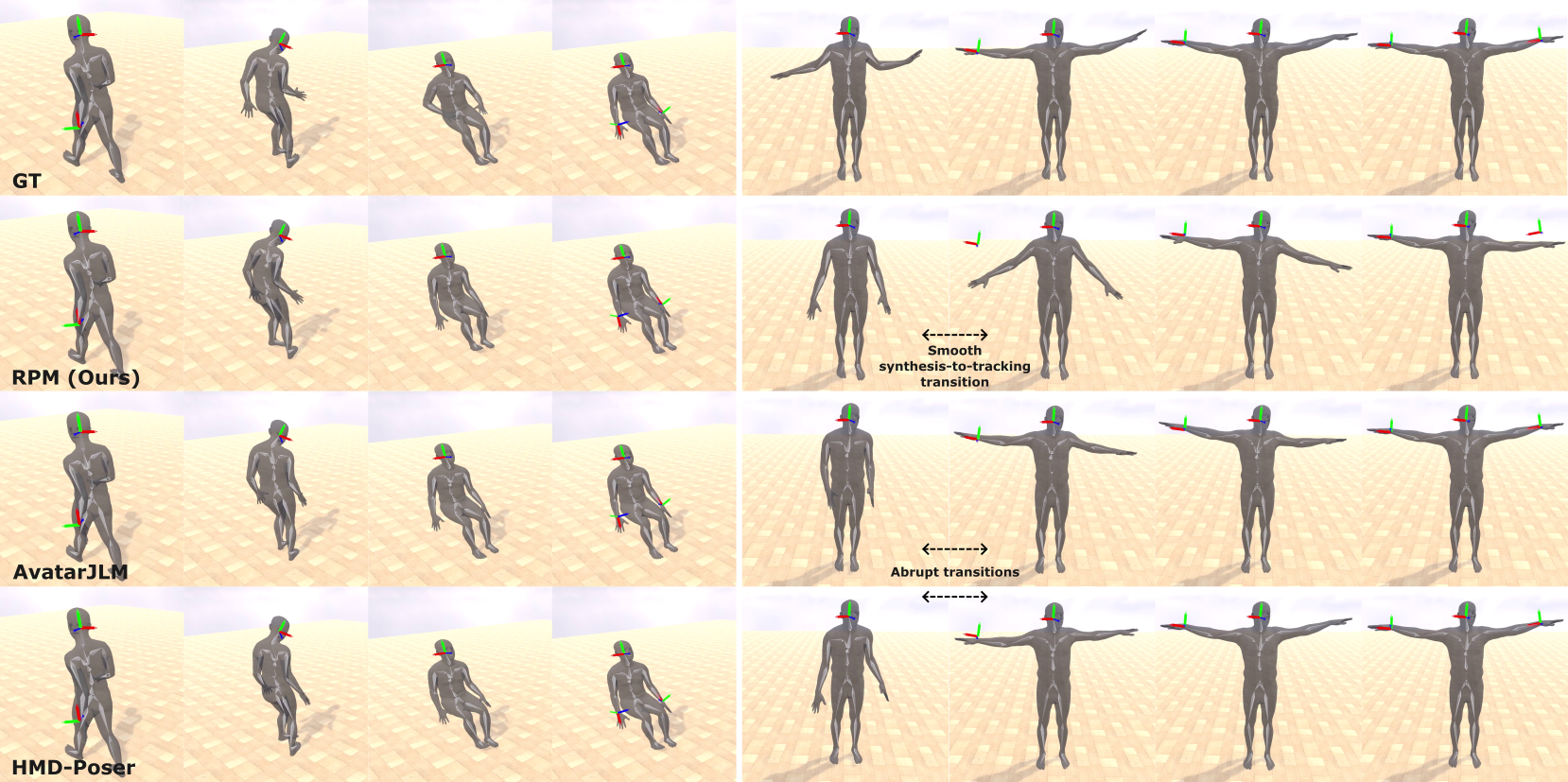}
    \caption{\textbf{Qualitative comparison on synthetic HT inputs (A-P1).} On the left, we show how RPM performs similarly to other state-of-the-art methods when the tracking inputs contain strong information on the full-body pose. However, more ambiguous input configurations might lead to wrong generated poses, as shown in the first column on the right example. When the tracking is recovered, \modelname{} is the only method that generates a smooth and realistic transition towards matching the new input.}
    \label{fig:qualitative_results_amass}
\end{figure*}

\begin{figure*}
    \centering
    \includegraphics[width=\linewidth]{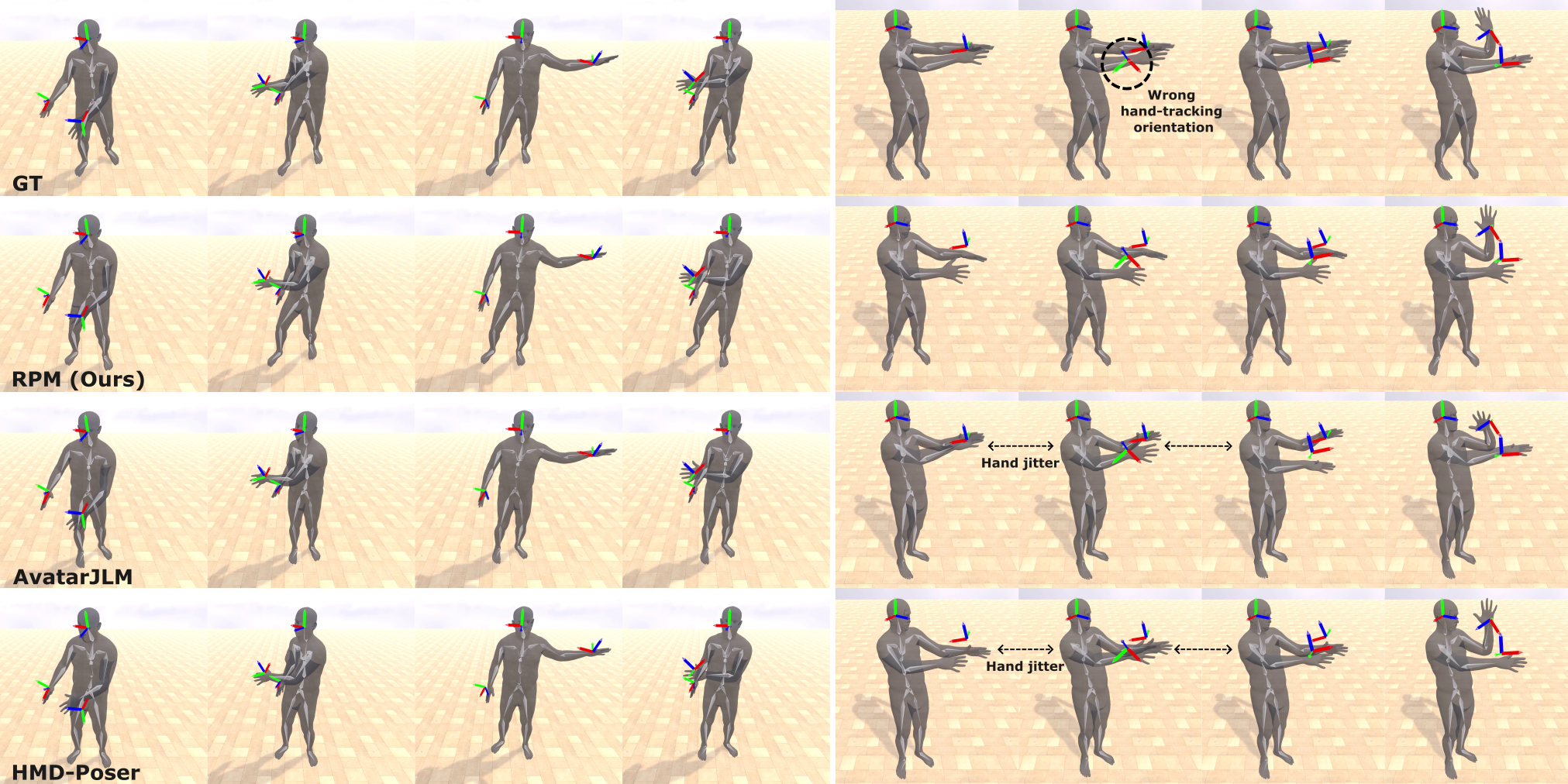}
    \caption{\textbf{Qualitative comparison on real inputs (GORP).} We show how \modelname{} is able to generate accurate and expressive full-body motion in highly dynamic sequences (left example, MC inputs). On the much more challenging HT setup (right), our method generates motion with less jitter when hand-tracking inputs are missing or noisy.}
    \label{fig:qualitative_results_gorp}
\end{figure*}

In \autoref{fig:qualitative_results_amass} and \ref{fig:qualitative_results_gorp}, we visually compare motions generated on the A-P1 and GORP datasets.  
We observe how \modelname{} synthesizes plausible motion during tracking signal losses.
Once the tracking is recovered, \modelname{} is the only method that generates a smooth transition towards matching it again. Other methods, instead, instantly snap the hands to the new tracking signal, breaking the continuity of the previous motion. The smoother transitions from \modelname{} lead to more realistic motion than other methods. 
\autoref{fig:trajectory_prediction} shows \modelname{}'s predicted motion during fast dynamic motion, and periods of tracking signal loss.
See \supp{}
for more details.


\section{Conclusions}
\label{sec:conclusions}

We introduced \modelname{}, a new architecture that runs in real-time and addresses, for the first time, the specific challenges of generating full-body motion from spatially and temporally sparse tracking inputs. \modelname{} generates smooth human motions even when tracking inputs are suddenly lost
, or recovered. Thanks to PCAF, we can control the model reactiveness to such tracking inputs, matching the needs of different applications. We also presented GORP, the first motion dataset with paired real VR tracking data. With it, we present the most complete benchmark up to date for this task. Our results show the performance gap that methods trained on synthetic datasets face when applied on real data, stressing \datasetname{}'s potential to boost progress in this field.

\textbf{Limitations and future work.}
Given the determinism of \modelname{}, generated motions sometimes lose expressivity during long periods of tracking signal loss.
Future work includes extending \modelname{} to refine multiple possible future motions and using other motion representations like~\cite{jiang2025egoposer}.

\textbf{Acknowledgements. }
This work has been partially supported by the Spanish project PID2022-136436NB-I00 and by ICREA under the ICREA Academia programme.

{
    \small
    \bibliographystyle{ieeenat_fullname}
    \bibliography{main}
}


\renewcommand*{\thesection}{\Alph{section}}
\renewcommand*{\thefigure}{\Alph{figure}}
\renewcommand*{\thetable}{\Alph{table}}
\renewcommand*{\theequation}{\Alph{equation}}
\setcounter{section}{0}
\setcounter{table}{0}
\setcounter{equation}{0}
\setcounter{figure}{0}

\clearpage
\maketitlesupplementary

This supplementary material is structured as follows. In \autoref{supp:gorp_dataset_details}, we provide additional details on the protocol followed to collect and process the new GORP dataset. Then, we describe all the implementation details of our Rolling Prediction Model (RPM) in \autoref{supp:implementation_details}. We also give more insights on both the synthetic and real benchmarks in \autoref{supp:benchmarks_details}. In \autoref{supp:freerunning} and \ref{supp:uncertainty}, we show quantitative studies on the effects of modifying the free-running length, and the uncertainty function, respectively.
In \autoref{supp:iterative}, we further discuss why RPM is not fed with the previous prediction, and provide additional experiments supporting our choice. 
In \autoref{sec:acc_smooth_tradeoff} we include additional experiments showing how RPM provides the best accuracy-smoothness trade-off.
In \autoref{sec:over_time}, we analyze RPM's robustness over long runs.
In \autoref{sec:scaling_function}, we support experimentally the choice of the hyperbolic tangent as the scaling function for PCAF.
Finally, we show and discuss additional qualitative comparisons with the state of the art in \autoref{supp:more_qualitative_results}.

\section{\datasetname{} dataset}
\label{supp:gorp_dataset_details}

\begin{figure}[t!]
    \centering
    \includegraphics[width=\linewidth]{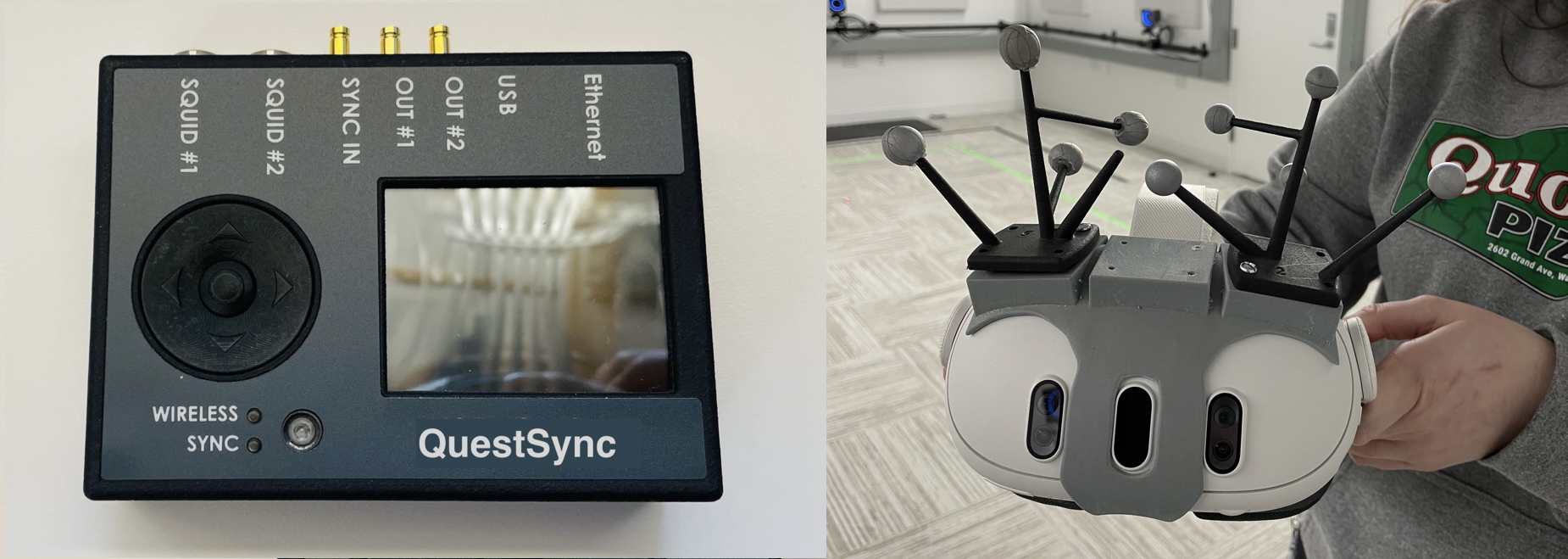}
    \caption{Left: the QuestSync device. Right: Meta Quest 3 with two marker trees attached.}
    \label{fig:hardware}
\end{figure}

\textbf{Data collection setup.}
The hardware setup includes a Meta Quest 3 headset, an Optitrack motion capture system with an eSync2 unit~\cite{esync} for synchronization, and a proprietary synchronization device, \textit{QuestSync}, as shown in \autoref{fig:hardware}. The QuestSync triggers data capture for both systems, and provides timestamps from a shared timeline to align the captured data. We use proprietary  software to synchronize QuestSync timestamps with Quest 3 data, such as headset, controller, and hand poses, which are already synchronized internally. 
For the Optirack system, QuestSync connects to eSync2 with a DIN-to-BNC cable, and provides it with a $30$-Hz SMTPE timecode~\cite{timecode}. The eSync2 can further subdivide this $30$ Hz signal to $120$ Hz for the Optitrack cameras. Since both the Quest 3 cameras and the Optitrack cameras rely on Infrared (IR) light for blob tracking, the QuestSync offsets trigger signals to eSync2 to prevent IR lights from Optitrack from interfering with the Quest 3. 

Both the Quest 3 and Optitrack stream data in real time to a PC that runs a proprietary data collection software. The Quest 3 streams tracking data through the Android Debug Bridge (ADB) and Meta Quest Link~\cite{link}. Optitrack streams 3D marker data through NatNet SDK~\cite{natnet}. These data streams include accurate capture timestamps from QuestSync in their protocol, so transmission and software latency are not a concern.

The Quest 3 already calibrates the headset tracking data with controller tracking data and hand tracking data internally. With synchronized marker data from Optitrack, we want to transform all Quest 3 data into the Optitrack space. To do so, we attach a marker tree on the Quest 3 headset as an Optitrack rigid body during data capture. The rigid transformation of the marker tree in Optitrack space, $\mathbf{M}^{MT}_t$, should represent the same motion as the Quest 3 headset tracking output in the Quest 3 space, $\mathbf{M}^{VR}_t$. We compute the constant transformation $\mathbf{T}$ from the Quest 3 space to the Optitrack space using \autoref{eqn:calibration} for a capture sequence of $N$ frames, where $\mathbf{O}$ represents the constant offset of the headset in the marker tree rigid body space.
\begin{equation}
     \mathbf{T}^*, \mathbf{O}^*= \operatorname*{argmin}_{\mathbf{T}, \mathbf{O}}\sum_{t=0}^N\|(\mathbf{T} \cdot\mathbf{M}^{VR}_t) - (\mathbf{M}^{MT}_t \cdot \mathbf{O})\|^2
    \label{eqn:calibration}
\end{equation}
Thanks to the accurate data synchronization, we achieve a calibration error of less than one millimeter on average across the entire dataset. 

\textbf{Data streams.} We retrieve a rich set of egocentric tracking signals from Quest 3. For both the headset and the controllers, it provides a rigid transformation, linear and angular velocities, and linear and angular accelerations at every time step, all synchronized and calibrated to the same space. The rigid transformation is computed through Visual Inertial Odometry (VIO) from stereo images and Inertial Measurement Units (IMUs) at $30$ Hz. The velocities and accelerations are computed from high-frequency IMUs and then downsampled to $30$ Hz, so they are much more accurate than finite differences of the positional data. In addition, each device has a flag to indicate whether its data is valid. This flag is always true for the headset in our dataset, but it may be false for the controllers when they are outside of the field of view of the headset cameras for an extended period. The origin of the headset is at the center between the eyes, and the origins of the controllers are somewhere between the thumb and the index finger, depending on how a user holds them.

Hand tracking data is exclusive to controller data in our dataset because it was collected before Multimodal mode~\cite{multimodal} was available on the Quest 3. The hands are tracked from the headset cameras~\cite{han2022umetrack} at $30$ Hz. For the purpose of our project, we captured only the rigid transformations of the wrists, as well as tracking confidence of each hand.

For body motion ground truth, we use SOMA and Mosh++~\cite{SOMA:ICCV:2021, AMASS} to fit SMPL body parameters to the Optitrack markers.

\begin{figure*}
    \centering
    \includegraphics[width=\linewidth]{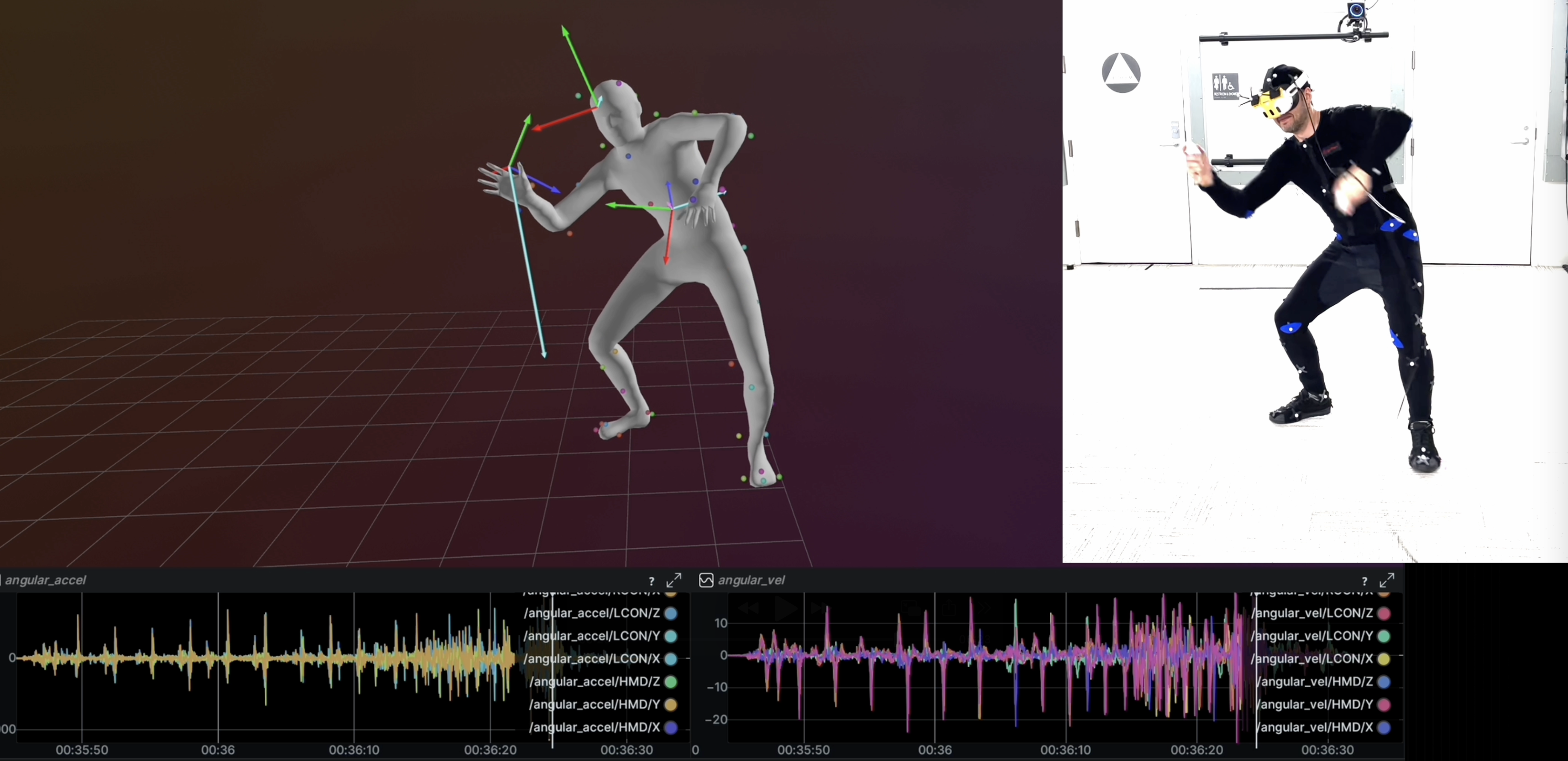}
    \caption{We capture both tracking signals from the Quest 3 headset and ground truth body motions from Optitrack. The Quest 3 tracking signal includes transformations of the headset and the controllers (or the wrists -- not shown here), linear velocities (purple arrows), linear accelerations (cyan arrows), angular velocity (bottom right), and angular accelerations (bottom left). The velocities and accelerations are from high-frequency IMU data. Ground truth body motions are solved from labeled 3D marker data (colored points).}
    \label{fig:gorp}
\end{figure*}

\textbf{Data protocol.} Our dataset consists of $28$ participants playing VR games using controllers and hands for 15 minutes respectively. We recruit participants from diverse demographics with various VR experience levels. There are $17$ male participants and $11$ female participants, ranging from $20$ to $65$ years old, 
who have played in VR for $0$ to $10+$ times. In a capture session, a participant wears a tight suit with reflective markers attached and puts on the Quest 3 headset, as shown in \autoref{fig:gorp}. After they feel comfortable in the VR environment, they are asked to choose one of two controller-based games, Beat Saber~\cite{beatsaber} or Fruit Ninja~\cite{fruitninja}. They can choose the game settings freely and start playing. After about 15 minutes or when they want to stop, they will take a 5 minutes break. Next, they are asked to choose one of two hand tracking based games, Hand Physics Lab~\cite{handslab} or Rogue Ascent VR~\cite{rogue}. They then play in the settings they choose for another 15 minutes or until they want to stop. They can also break or stop any time during the session. In total, we collected about 17 hours of realistic VR gameplay data with ground truth body motions. 


\section{Implementation details}
\label{supp:implementation_details}

\textbf{Architecture details.} The design of our model, $f_\theta$, is inspired by \cite{tevet2023human}. First, we concatenate the sequence of past motion $\X_t$ with a sequence of $W$ blank pose tokens (i.e., all zeros), which are passed through a linear layer and merged with a sinusoidal positional encoding (PE). 
A cross-attention layer then uses these tokens as queries to attend to the past and current tracking inputs, $\C_t$, which are fed as keys and values after adding a sinusoidal PE. The result is processed by four Transformer encoder layers~\cite{vaswani2017attention} with GeLU activation functions~\cite{hendrycks2016gaussian}. The first $M$ output feature vectors are discarded, retaining only the last $W$. Each attention layer incorporates layer normalization and a residual connection. Finally, a linear layer transforms the output features into the predicted poses, $f_{\theta}(\X_t, \C_t)$, which are then used by our Prediction Consistency Anchor Function (PCAF) to update the previous prediction, $P_{t-1}$, into $P_{t}$. The dimension of the Transformer latent space is set to 512. Feature vectors have a dimension of 132 for poses in $\X_t$ and $\P_t$, and 54 for tracking inputs in $\C_t$.

\textbf{Further implementation details. }
We optimize \modelname{} using Adam~\cite{kingma2014adam} and train it for 80k iterations with a batch size of 512. The initial learning rate is set to 3e-4, which decreases to 3e-5 after 50k iterations. Weight decay is set to 1e-4. The weights for all losses in Eq. 3 are set to 1. During training, we simulate tracking signal losses by dropping a segment of tracking inputs of length  $L \sim \mathcal{U}(1, I{+}1{+}\text{FR})$ with a 10\% probability, where $I$ is the number of past tracking inputs used, and FR the maximum free-running length. For all baselines, $L \sim \mathcal{U}(1, I_{\text{baseline}})$, where $I_{\text{baseline}}$ is the length of the tracking input sequence used by each baseline. For the AGRoL baseline~\cite{du2023agrol}, we generate motion for the first 196 frames offline, as proposed in the original implementation. We then use this output to autoregressively inpaint a new pose at the end. For \modelname{}, both the motion context and the initial prediction are initialized with ground truth poses. To avoid evaluating during this period, the first second of all sequences is discarded.

\section{Benchmarks details}
\label{supp:benchmarks_details}

\textbf{Synthetic inputs: AMASS (A-P1 and A-P2).} We build the motion controllers (MC) synthetic setup on AMASS in the same way as prior works~\cite{jiang2022avatarposer, du2023agrol}, that is, taking the head and wrists joint positions as tracking inputs. For the hand-tracking (HT) setup, we simulate gaps in the tracking inputs. Specifically, for each wrist in each motion sequence, we simulate periods of tracking signal loss with length  $L\sim \mathcal{U}(0.5, 2)$ seconds, with a 2.5\% probability of starting at a given timestep. We also enforce a distance of 2 s between consecutive gaps so we can measure the models' ability to recover from a tracking signal loss.

\textbf{Real inputs: \datasetname{}.} The GORP dataset was randomly split into subject-independent training and test sets, ensuring an 80/20 proportion of total duration, respectively. All models were trained using both sequences with real MC or HT sensing signals.
We approximated the 6-DOF of the wrists and the SMPL head by adding a fixed offset to the motion controllers and headset position. 
The confidence threshold of the hand-tracking system was set to 0.8, resulting in 3.16/1.50\% of missing left/right hands in the training set and 4.05/1.49\% in the test set. The average length of tracking signal loss segments is 0.84s for the training set and 0.77s for the test set.

\section{Analysis of the free-running length}
\label{supp:freerunning}

\begin{figure}[t!]
    \centering
    \includegraphics[width=\linewidth]{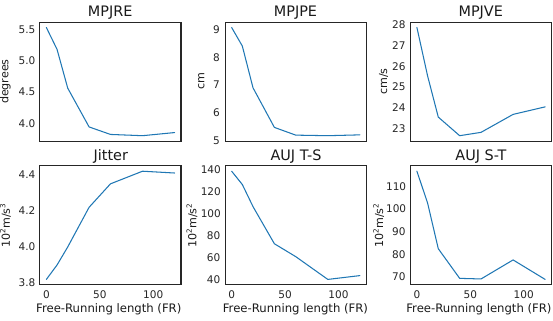}
    \vspace{-6mm}
    \caption{\textbf{Free-running length.} We observe that the free-running stage during \modelname{}'s training is essential for making the network robust to mismatches between tracking inputs ($\C_t$) and previously generated motion ($\X_t$). Both accuracy (MPJRE, MPJPE) and smoothness during transitions (AUJ T-S/S-T) improve significantly with longer free-running periods.}
    \label{fig:fr_study}
\end{figure}

In \autoref{fig:fr_study}, we show the effects of the free-running length on the accuracy and smoothness of the generated motion for A-P1. Overall, we observe that longer free-running periods result in higher accuracy (i.e., lower MPJRE and MPJPE), with values converging to the best performance at around 50 frames (0.83s). Regarding smoothness, longer FR values cause a slight increase in jitter (from 3.8 to 4.4) and a reduction in the AUJ for both types of transition. We hypothesize that the improved smoothness during transitions is due to a better simulation at training time of conditions encountered during inference, such as mismatches between motion context and tracking inputs after periods of tracking signal loss (see Fig. 3, right). The best smoothness values during transitions are achieved at around FR=90 frames (1.5s). For all our experiments, we use FR=60 frames (1s), as it offers a good balance between accuracy, smoothness, and training efficiency.

 
\section{Analysis of the uncertainty function}
\label{supp:uncertainty}

\begin{figure}[t!]
    \centering
    \includegraphics[width=0.8\linewidth]{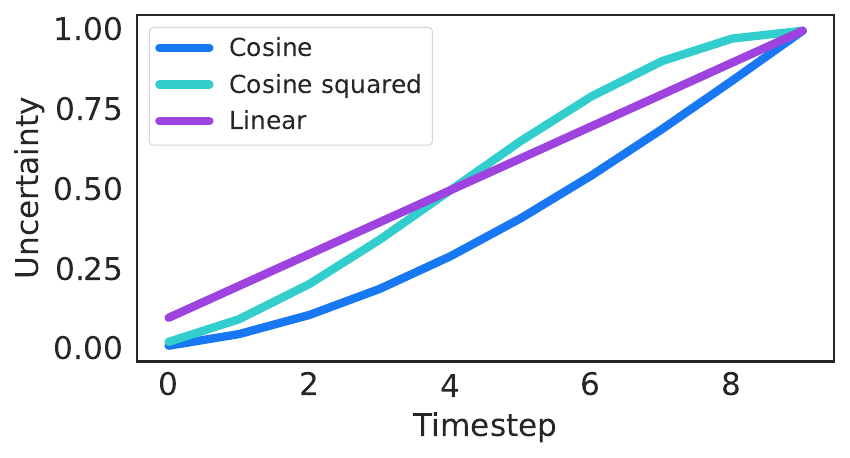}
    \caption{\textbf{Visualization of the uncertainty functions.} We explored three uncertainty functions: cosine (\autoref{eq:cosine}), cosine squared (\autoref{eq:cosine_sq}), and linear (\autoref{eq:linear}). In all cases, the uncertainty level increases with the distance from the present.}
    \label{fig:uncertainty_functions}
\end{figure}

In Sec. 3.2, we introduce the PCAF reparameterization (Eq. 2), which incorporates an uncertainty function to regulate the correction magnitude allowed at each future time horizon. The objective is to model the increasing uncertainty of future motion as the prediction extends further into the future. Consequently, the corrections applied to previous predictions should scale proportionally with this uncertainty. We explored three different uncertainty functions: cosine (\autoref{eq:cosine}), cosine squared (\autoref{eq:cosine_sq}), and linear (\autoref{eq:linear}):

\begin{equation}
    f_{\text{cos}}(\tau) = 1- \cos \Big( \frac{\tau+1}{W} \cdot \frac{\pi}{2}\Big),
    \label{eq:cosine}
\end{equation}

\begin{equation}
    f_{\text{cos sq.}}(\tau) = 1- \cos \Big( \frac{\tau+1}{W} \cdot \frac{\pi}{2}\Big)^2,
    \label{eq:cosine_sq}
\end{equation}

\begin{equation}
    f_{\text{lin}}(\tau) = \frac{\tau+1}{W},
    \label{eq:linear}
\end{equation}

\noindent where $\tau \in [0, W-1]$ represents the distance to the present, in frames. A visualization of the three functions with $W{=}10$ is provided in \autoref{fig:uncertainty_functions}. The results of \modelname{} trained on A-P1 are presented in \autoref{tab:study_schedule}. While accuracy metrics are comparable across the three functions, the cosine function produces smoother motion during transitions (as shown by lower \aujts{} and \aujst{}). Given the importance of smooth transitions for the problem we tackle in this work, we selected the cosine function as the uncertainty function for all our experiments.

\begin{table}[t!]
    \footnotesize	
    \addtolength{\tabcolsep}{-0.35em}
    \centering
    \begin{tabular}{lllllll}
    \toprule
    Uncertainty (U) & MPJRE & MPJPE & MPJVE & Jitter & \aujts & \aujst \\
    \midrule
    Cosine & $\textbf{3.82}$ & $\textbf{5.18}$ & $22.83$ & $4.35$ & $\textbf{60.51}$ & $\textbf{69.02}$ \\
    Cosine squared & $3.92$ & $5.37$ & $22.51$ & $4.28$ & $60.85$ & $78.97$ \\
    Linear & $3.92$ & $5.33$ & $\textbf{22.05}$ & $\textbf{4.25}$ & $71.55$ & $80.47$ \\
    \bottomrule
    \end{tabular}
    
    \caption{\textbf{Comparison of different uncertainty functions.}
    We observe that modifying the uncertainty function of PCAF primarily impacts the smoothness during synthesis-to-tracking (\aujst) and tracking-to-synthesis (\aujts) transitions.}
    \label{tab:study_schedule}
\end{table}


\section{On the iterative prediction refinement}
\label{supp:iterative}

\modelname{} uses the PCAF reparameterization to refine the previous prediction, $\P_{t-1}$. However, this previous prediction is not fed directly to $f_\theta$, which only observes the previously generated motion, $\X_t$, and the history of tracking inputs, $\C_t$. At the end of Sec. 3.2, we note that $\X_t$ serves as a proxy for the previous prediction $\P_{t-1}$ under the deterministic paradigm, which assumes that only a single future is predicted, as opposed to a multimodal distribution (i.e., stochastic prediction). To validate this claim, we train a version of \modelname{} that is explicitly fed with $\P_{t-1}$, thereby providing direct access to the previous prediction. 
The results, shown in \autoref{tab:prevpred}, support our hypothesis: explicit access to $\P_{t-1}$ neither improves accuracy nor generates smoother results. 
We also train another version where, instead of feeding the previous prediction, we provide a noisy version of it to simulate the uncertainty of the future as a decreasing signal-to-noise ratio, inspired by rolling diffusion models~\cite{ruhe2024rolling, zhang2023tedi}. Similarly, no significant improvement is observed over our simpler definition. Nonetheless, we hypothesize that these alternative definitions could be beneficial if \modelname{} were extended to refine stochastic human motion predictions, where previous predictions may not be directly inferable from $\X_t$ due to their multimodal nature.

\begin{table}[]
    \footnotesize	
    \addtolength{\tabcolsep}{-0.4em}
    \centering
    \begin{tabular}{lllllll}
    \toprule
    Forward function & MPJRE & MPJPE & MPJVE & Jitter & \aujts & \aujst \\
    \midrule
    $f_\theta (\X_t , \C_t)$ & $3.82$ & $5.18$ & $22.83$ & $4.35$ & $60.51$ & $69.02$ \\
    $f_\theta (\X_t , \C_t, \P_{t-1})$ & $3.80$ & $5.19$ & $\textbf{22.47}$ & $\textbf{4.12}$ & $82.75$ & $\textbf{65.22}$ \\
    $f_\theta (\X_t , \C_t, \P_{t-1}^{\text{noisy}})$ & $\textbf{3.78}$ & $\textbf{5.13}$ & $22.97$ & $5.40$ & $55.74$ & $103.23$ \\
    \bottomrule
    \end{tabular}
    \caption{Study on different techniques to refine the rolling prediction with \modelname{}. We observe that explicitly feeding the previous prediction ($\P_{t-1}$) leads to results similar to those of our simpler model (1st row). Similarly, simulating the uncertainty of the previous prediction by adding noise with increasing variance to it ($\P^{\text{noisy}}_{t-1}$), as proposed in \cite{ruhe2024rolling}, does not lead to a better performance than \modelname{} either.}
    \label{tab:prevpred}
\end{table}

\section{Accuracy-smoothness trade-off}
\label{sec:acc_smooth_tradeoff}

\begin{figure*}[t]
    \centering
        \includegraphics[width=\linewidth]{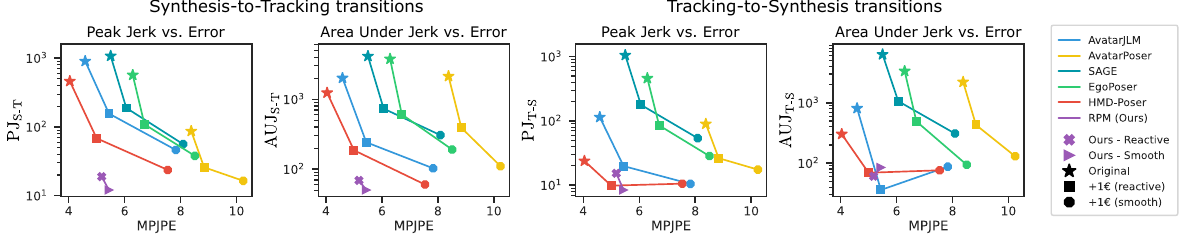}
        \caption{\textbf{Accuracy-smoothness trade-off on A-P1-HT.} Baselines require significant accuracy loss to achieve competitive smoothness when combined with traditional low-pass filters. On synthesis-to-tracking transitions, RPM achieves the best balance to date, positioning itself in the lower-left of the plot and surpassing the previous Pareto frontier set by HMD-Poser. At the same time, RPM attains a similar trade-off as HMD-Poser and AvatarJLM on tracking-to-synthesis transitions.}
        \label{fig:pareto}
\end{figure*}

In Sec. 4, we discussed how current state-of-the-art methods lag behind RPM in terms of smoothness and, consequently, motion realism. In return, our model sacrifices some accuracy. One might think that state-of-the-art methods could also produce such smooth results if they were willing to sacrifice accuracy, for instance, by incorporating a low-pass filter. To test this hypothesis, we applied a 1€ filter~\cite{casiez20121} to the output of our baselines. The filter parameters were optimized through grid search to maximize the smoothness metrics. However, we found that achieving competitive smoothness introduced excessive latency to the generated motion, significantly reducing accuracy. To further explore this accuracy-smoothness trade-off, we identified another set of filter parameters offering the best balance between accuracy and smoothness. We refer to these two sets of parameters as smooth and reactive, respectively. Results in \autoref{tab:eurofilter} demonstrate that all baselines still produce motion with more discontinuities than RPM when using a mild low-pass filter that minimally affects accuracy. When targeting smoothness levels comparable to RPM for synthesis-to-tracking transitions, their accuracy drops by more than 50\%, placing them far behind RPM in terms of accuracy. We illustrate this trade-off in \autoref{fig:pareto}. Our model provides the optimal accuracy-smoothness trade-off, particularly during synthesis-to-tracking transitions, which was the primary goal of our work.

\begin{table*}[t!]
    \centering
    \footnotesize	
    \vspace{-3mm}
    \addtolength{\tabcolsep}{-0.3em}
    \begin{tabular}{lcccccccc}
    \toprule
    Model & MPJRE & MPJPE & MPJVE & Jitter & \pjts{} & \aujts{} & \pjst{} & \aujst{} \\
    \midrule
    AvatarPoser~\cite{jiang2022avatarposer} & $5.62$ & $8.38$ & $44.26$ & $26.10$ & $89.87$ & $2215.07$ & $86.93$ & $2133.20$ \\
    \hspace{0.5cm} + 1€ (reactive) & $5.77$ & $8.84$ & $40.45$ & $9.34$ & $26.30$ & $429.34$ & $25.93$ & $389.75$ \\
    \hspace{0.5cm} + 1€ (smooth) & $6.13$ & $10.24$ & $43.69$ & $6.21$ & $17.46$ & $129.79$ & $16.48$ & $110.13$ \\
    \midrule
    EgoPoser~\cite{jiang2025egoposer} & $4.61$ & $6.29$ & $42.91$ & $29.15$ & $461.18$ & $3356.38$ & $567.42$ & $3773.29$ \\
    \hspace{0.5cm} + 1€ (reactive) & $4.75$ & $6.71$ & $34.63$ & $7.57$ & $85.27$ & $491.61$ & $108.95$ & $606.55$ \\
    \hspace{0.5cm} + 1€ (smooth) & $5.26$ & $8.51$ & $40.44$ & $4.89$ & $28.37$ & $93.64$ & $38.00$ & $190.45$ \\
    \midrule
    SAGE~\cite{feng2024sage} & $4.21$ & $5.50$ & $46.38$ & $33.55$ & $1056.33$ & $6337.79$ & $1073.65$ & $4131.33$ \\
    \hspace{0.5cm} + 1€ (reactive) & $4.40$ & $6.06$ & $35.90$ & $9.82$ & $178.16$ & $1064.44$ & $188.51$ & $733.46$ \\
    \hspace{0.5cm} + 1€ (smooth) & $4.99$ & $8.09$ & $41.22$ & $6.28$ & $53.90$ & $308.76$ & $56.25$ & $306.36$ \\
    \midrule
    AvatarJLM~\cite{zheng2023avatarjlm} & $4.18$ & $4.59$ & $27.30$ & $12.79$ & $114.36$ & $811.13$ & $901.75$ & $2008.13$ \\
    \hspace{0.5cm} + 1€ (reactive) & $4.44$ & $5.44$ & $29.35$ & $4.67$ & $19.82$ & $\textbf{35.86}$ & $156.25$ & $242.25$ \\
    \hspace{0.5cm} + 1€ (smooth) & $5.07$ & $7.83$ & $39.66$ & $4.03$ & $10.47$ & $87.31$ & $46.31$ & $102.73$ \\
    \midrule
    HMD-Poser~\cite{dai2024hmdposer} & $\textbf{3.34}$ & $\textbf{4.04}$ & $\textbf{22.34}$ & $7.35$ & $23.84$ & $302.58$ & $461.91$ & $1236.47$ \\
    \hspace{0.5cm} + 1€ (reactive) & $3.71$ & $5.00$ & $27.18$ & $\textbf{3.98}$ & $9.83$ & $69.17$ & $68.38$ & $186.21$ \\
    \hspace{0.5cm} + 1€ (smooth) & $4.55$ & $7.54$ & $39.06$ & $4.10$ & $10.55$ & $75.98$ & $23.75$ & $60.14$ \\
    \midrule
    RPM - Reactive & $3.82$ & $5.18$ & $\textbf{22.83}$ & $4.35$ & $15.28$ & $60.51$ & $18.98$ & $69.02$ \\
    RPM - Smooth & $3.98$ & $5.44$ & $24.04$ & $4.29$ & $\textbf{8.41}$ & $84.81$ & $\textbf{12.12}$ & $\textbf{50.23}$ \\
    \bottomrule
    \end{tabular}
\caption{\textbf{Comparison of RPM with baselines using 1€ filters on A-P1-HT.} Applying both versions of the 1€ filter (reactive and smooth versions, see~\autoref{sec:acc_smooth_tradeoff}) to baselines improves smoothness but significantly reduces accuracy because of latency. To match RPM’s smoothness levels (\aujst{}), baselines must sacrifice over 50\% of their accuracy (MPJPE).}
\label{tab:eurofilter}
\end{table*}

\section{Performance over time}
\label{sec:over_time}

\begin{figure}[t]
    \centering
    \includegraphics[width=\linewidth]{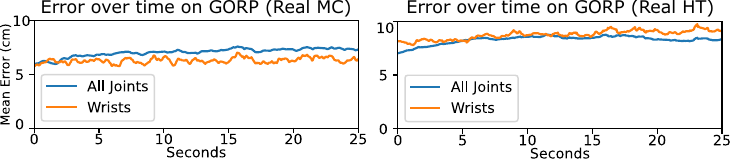}
    \caption{\textbf{Error evolution over time.} RPM’s error on both real MC/HT setups in the GORP dataset remains stable over time, demonstrating its robustness in long executions. Wrist error also remains stable and is comparable in magnitude to the overall error.}
    \label{fig:degeneration}
\end{figure}

Autoregressive methods may experience degeneration issues over time. To showcase RPM’s robustness and its potential for real-world deployment, we plot the error evolution over 25 seconds in the real MC/HT setups using the GORP dataset in \autoref{fig:degeneration}. We observe that the error remains stable in both setups. Additionally, the wrist error does not escalate and stays close to the overall error’s magnitude. Interestingly, despite the tracker position being available, the network still struggles to match it closely. This issue is commonly observed in methods synthesizing full-body motion from sparse tracking inputs \cite{jiang2024manikin}. We argue that this is expected in RPMs, as certain accuracy is traded off for smoother and more realistic motion, as discussed in \autoref{sec:acc_smooth_tradeoff}.

\section{Scaling function in PCAF}
\label{sec:scaling_function}

At the end of Sec. 3.2, we discussed the rationale behind selecting the hyperbolic tangent as the scaling function in PCAF. In this section, we present the corresponding experimental validation. We trained RPM with PCAF using three functions: the hyperbolic tangent, a scaled sigmoid ($2*sigmoid(...) - 1$), and a linear function constrained between -1 and 1 (referred to as \textit{straight}). As shown in \autoref{tab:tanh_pred10}, the model trained with the hyperbolic tangent produces the smoothest synthesis-to-tracking transitions. The \textit{straight} function achieves similar accuracy to the hyperbolic tangent, but with slightly more abrupt transitions. In contrast, the sigmoid function leads to transitions that are too abrupt. We hypothesize that the sigmoid function generates large gradients when substantial correction is needed, promoting strong corrections during prediction refinement rather than improving long-term predictions. This issue does not occur with the other two functions. We selected the hyperbolic tangent for its continuity and differentiability.

\begin{table}[t]
    \addtolength{\tabcolsep}{-0.1em}
    \footnotesize	
    \centering
    \begin{tabular}{llllll}
    \toprule
    & MPJPE & MPJVE & Jitter & \aujts{} & \aujst{} \\
    \midrule
    RPM - Reactive \\
    \midrule
    Tanh & $5.18$ & $22.83$ & $4.35$ & $60.51$ & $\textbf{69.02}$ \\
    Sigmoid & $5.21$ & $23.45$ & $5.14$ & $\textbf{55.11}$ & $137.88$ \\
    Straight & $\textbf{5.11}$ & $\textbf{22.58}$ & $\textbf{4.30}$ & $61.55$ & $71.19$ \\
    \midrule
    RPM - Smooth \\
    \midrule
    Tanh & $\textbf{5.59}$ & $23.80$ & $\textbf{5.05}$ & $58.70$ & $\textbf{175.86}$ \\
    Sigmoid & $5.63$ & $25.32$ & $6.76$ & $232.50$ & $431.62$ \\
    Straight & $\textbf{5.59}$ & $\textbf{23.69}$ & $5.37$ & $\textbf{53.68}$ & $187.27$ \\
    \bottomrule
    \end{tabular}
    \caption{\textbf{Effect of PCAF scaling function iA-P1-HT.} We observe how the hyperbolic tangent and the straight functions provide the best accuracy and transitions smoothness.}
    \label{tab:tanh_pred10}
\end{table}


\section{Additional qualitative results}
\label{supp:more_qualitative_results}

\begin{figure*}[t!]
    \centering
    \includegraphics[width=\linewidth]{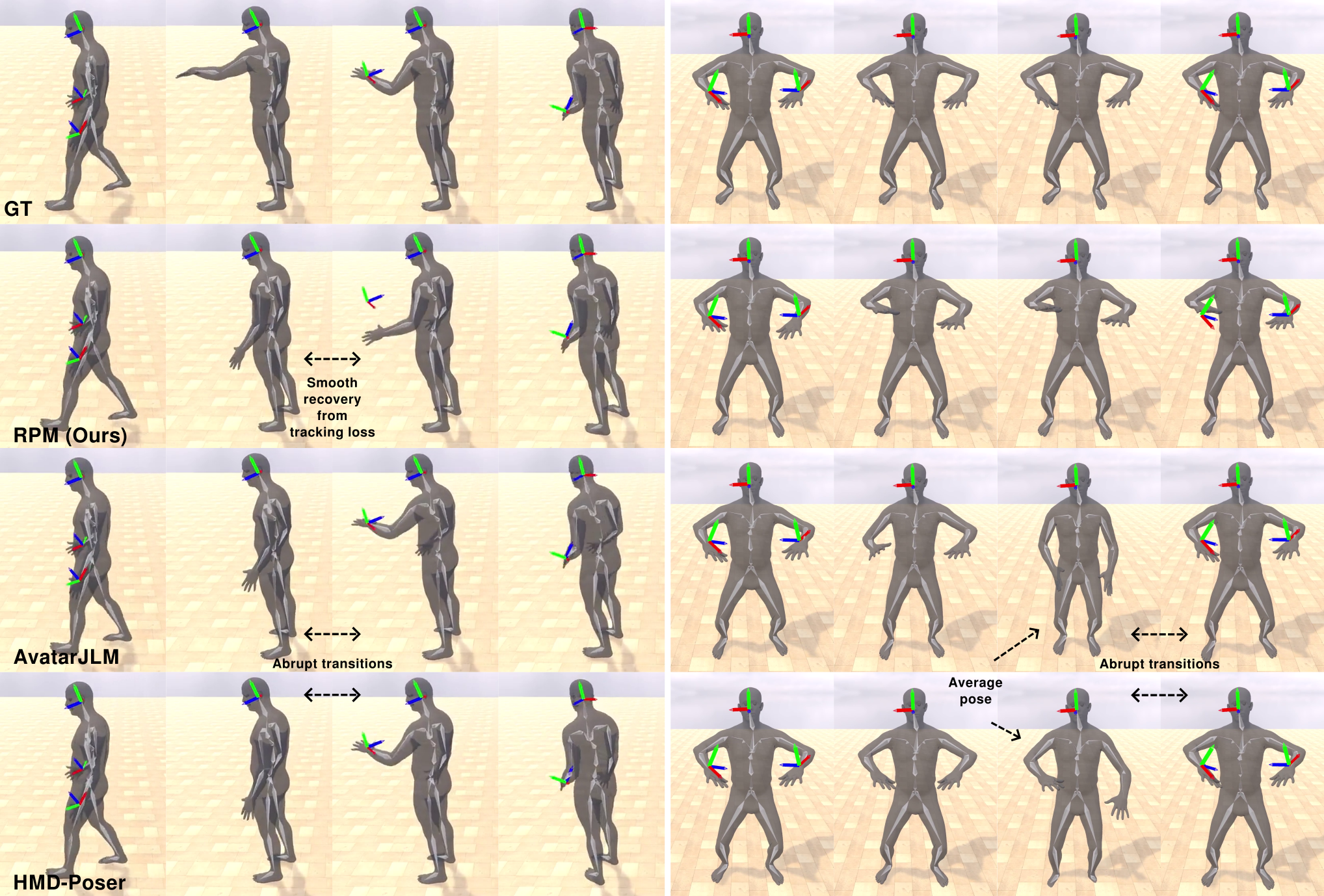}
    \vspace{-6mm}
    \caption{\textbf{More qualitative results on synthetic hand-tracking (HT) inputs (A-P1).} On the left, we observe that when the left-hand tracking is lost, all methods fail to generate the \textit{grabbing} action. Once the hand-tracking signal is recovered, AvatarJLM and HMD-Poser snap the hand abruptly to correct the mistake, whereas our method generates a smoother and more realistic transition to match the signal again. On the right, we present an example illustrating how state-of-the-art methods tend to generate an average pose as uncertainty increases during prolonged hand-tracking signal losses. In contrast, \modelname{} produces motion that remains coherent with the past context. }
    \label{fig:qualitative_results_supp_1}
\vspace{-3mm}
\end{figure*}

\begin{figure*}[t!]
    \centering
    \includegraphics[width=\linewidth]{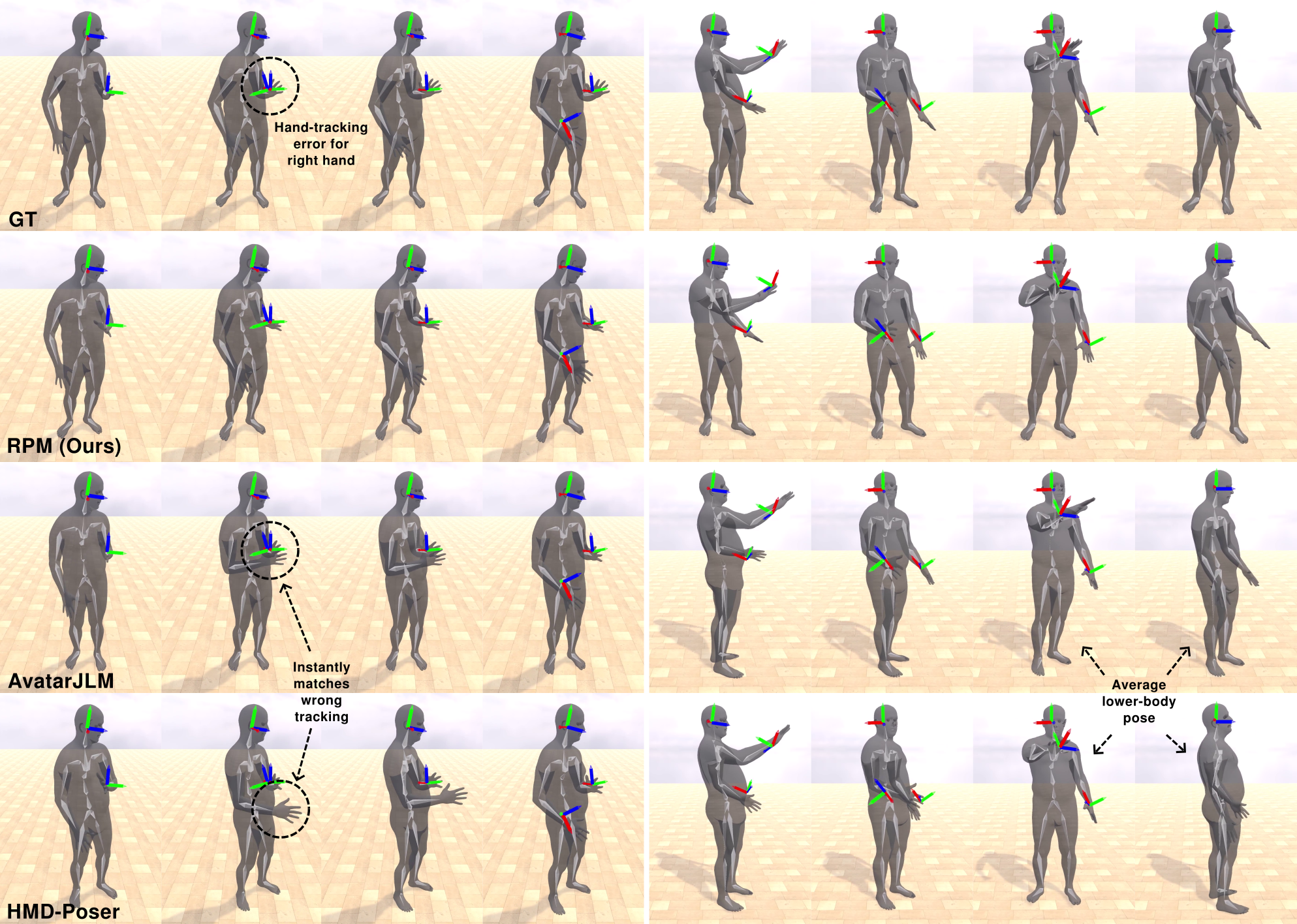}
    \vspace{-6mm}
    \caption{\textbf{More qualitative results on real hand-tracking (HT) inputs (GORP).} On the left, we observe that false-positive hand-tracking detections cause AvatarJLM and HMD-Poser to disrupt the continuity of the previous motion by abruptly aligning with the new tracking input. In contrast, \modelname{} generates motion that maintains coherence with prior predictions, thereby avoiding instantaneous hand snapping. On the right, we see that state-of-the-art methods often generate lower-body average poses, simply aligning their orientation with the tracking inputs. Instead, \modelname{} preserves coherence with the prior lower-body motion, producing more realistic full-body motion.}
    \label{fig:qualitative_results_supp_2}
\vspace{-3mm}
\end{figure*}

We provide additional visual comparisons in \autoref{fig:qualitative_results_supp_1} for A-P1 and in \autoref{fig:qualitative_results_supp_2} for GORP. In this section, we discuss several improvements that \modelname{} offers compared to the state of the art in terms of visual quality. We refer the reader to the attached videos, which showcase all the qualitative results from the main paper and supplementary material. Additionally, we include a demo video highlighting RPM's strengths in comparison to the state of the art.

\textbf{Synthesis-to-tracking transitioning.} These additional visual comparisons reinforce our observations from Sec. 4.3: only our method generates smooth and realistic synthesis-to-tracking transitions. While AvatarJLM and HMD-Poser repeatedly snap the hand to the recovered tracking signal after each loss, \modelname{} instead smoothly catches up with the tracking signal’s position and motion dynamics.

\textbf{Robustness to noise.} \modelname{} demonstrates greater robustness to noisy tracking inputs. By reformulating the pose generation task as a sequential refinement of previously predicted motion, the network learns to better handle sudden false hand-tracking detections (\autoref{fig:qualitative_results_supp_2}-left). We hypothesize that, at training time, these false positives fail to provide a signal capable of refining high-frequency motion details, which are incorporated during the last refining step before the pose generation. Instead, the network leverages such new signal to correct the low-frequency characteristics of the long-term predicted motion. However, these corrections do not persist under noisy hand-tracking inputs, as subsequent accurate hand-tracking inputs override them.

\textbf{Full-body pose coherence.} Lastly, our method maintains the coherence of the full-body pose along time. The main reason behind this is RPM's explicit conditioning on the previously generated motion ($\X_t$), instead of just the past and current tracking inputs ($\C_t$) as other methods do. In GORP, this ensures that dynamic upper-body motion leading to torso rotations does not result in unrealistic lower-body rotations (\autoref{fig:qualitative_results_supp_2}-right). Consequently, we observe more expressive lower-body motion, including stepping while turning. Conversely, in AvatarJLM and HMD-Poser, the tendency to align the lower-body with an average pose often causes severe foot sliding -- i.e., moving feet while standing as if the person floated. Notably, foot sliding remains a common issue across all methods for this task, including ours. Further research is needed in this area, as we hypothesize that the lack of body shape awareness and the reliance on headset-driven motion are potential reasons for it. We highlight the value of GORP as a benchmark for addressing this challenge, as its motion is predominantly in place, featuring frequent small in-place steps, knee torsion, and significant upper-body movement. These characteristics require networks to adapt the entire kinematic chain, including the lower-body and the spine, to minimize foot-sliding artifacts.

\end{document}